\documentclass[10pt,twocolumn,letterpaper]{article}

\usepackage{cvpr}              

\usepackage{graphicx}
\usepackage{amsmath}
\usepackage{amssymb}
\usepackage{booktabs}
\usepackage{verbatim}
\usepackage[ruled, lined, boxed]{algorithm2e}
\usepackage{multirow}
\usepackage{enumitem}
\usepackage{microtype}

\usepackage[pagebackref,breaklinks,colorlinks]{hyperref}

\usepackage[capitalize]{cleveref}
\crefname{section}{Sec.}{Secs.}
\Crefname{section}{Section}{Sections}
\Crefname{table}{Table}{Tables}
\crefname{table}{Tab.}{Tabs.}

\def\cvprPaperID{2384} 
\def\confName{CVPR}
\def\confYear{2022}

\begin{document}


\title{Few Shot Generative Model Adaption via Relaxed Spatial Structural Alignment}

\author{
Jiayu Xiao$^{1,2}$, Liang Li$^{1}$\thanks{Corresponding author.}, Chaofei Wang$^{3}$\thanks{Equal contribution.}, Zheng-Jun Zha$^{4}$, Qingming Huang$^{1,2,5}$~~~~~~~ \\
$^{1}$Key Lab of Intell. Info. Process., Inst. of Comput. Tech., CAS, Beijing, China~~~~~~~~~~~~~~\\
$^{2}$University of Chinese Academy of Sciences, Beijing, China, $^{3}$Department of Automation, Tsinghua University \\
$^{4}$University of Science and Technology of China, China, $^{5}$Peng Cheng Laboratory, Shenzhen, China~~~~~~\\
{\tt \small jiayu.xiao@vipl.ict.ac.cn,liang.li@ict.ac.cn,} \\
{\tt \small wangcf18@mails.tsinghua.edu.cn,zhazj@ustc.edu.cn,qmhuang@ucas.ac.cn}
}


\maketitle
\vspace{-0.4cm}
\begin{abstract}
\vspace{-0.2cm}
Training a generative adversarial network (GAN) with limited data has been a challenging task. A feasible solution is to start with a GAN well-trained on a large scale source domain and adapt it to the target domain with a few samples, termed as few shot generative model adaption. However, existing methods are prone to model overfitting and collapse in extremely few shot setting (less than 10). To solve this problem, we propose a relaxed spatial structural alignment (RSSA) method to calibrate the target generative models during the adaption. We design a cross-domain spatial structural consistency loss comprising the self-correlation and  disturbance correlation consistency loss. It helps align the spatial structural information between the synthesis image pairs of the source and target domains. To relax the cross-domain alignment, we compress the original latent space of generative models to a subspace. Image pairs generated from the subspace are pulled closer. Qualitative and quantitative experiments show that our method consistently surpasses the state-of-the-art methods in few shot setting. Our source code:
\url{https://github.com/StevenShaw1999/RSSA}. 





\end{abstract}

\vspace{-0.6cm}
\section{Introduction}
\label{sec:intro}
\begin{figure}[t]
  \centering
   \includegraphics[width=0.99\linewidth]{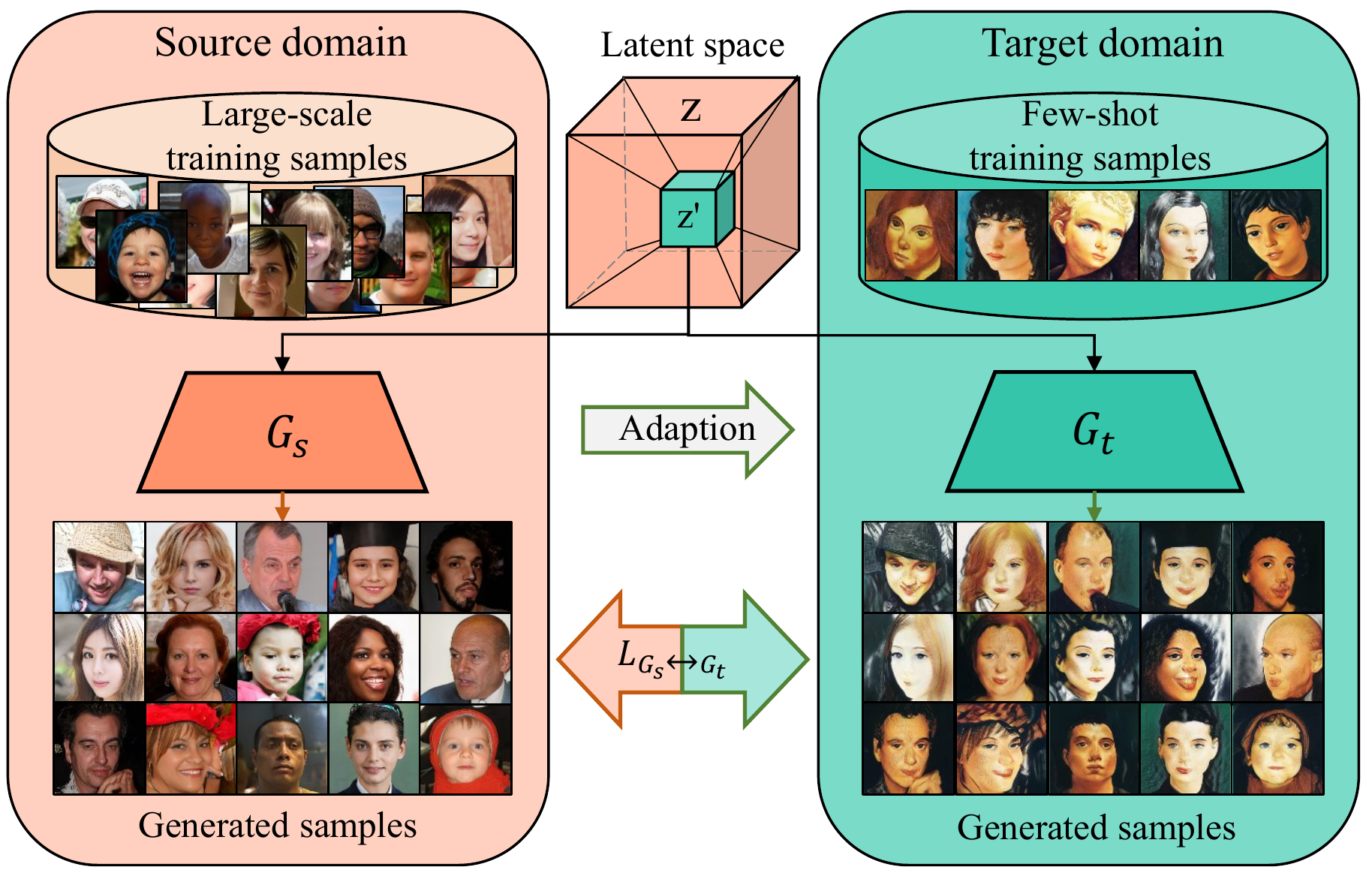}
    \vspace{-1.5mm}
   \caption{Few shot generative model adaption and our motivation.
   Start with a pre-trained model of the source domain $G_s$ and adapt to the target domain to get $G_t$
   by using extremely few (such as 5) training images of the target domain. We compress the latent space to a subspace close to target domain, and align the spatial structural information of synthesis pairs generated by $G_s$ and $G_t$.}
   \label{fig:setting}
   \vspace{-5.5mm}
\end{figure}
\vspace{-0.15cm}
Generative adversarial networks (GANs) have achieved promising results in various computer vision scenarios such as natural image synthesis \cite{brock2018large,karras2019style}, image to image translation\cite{zhu2017unpaired} and image inpainting\cite{yeh2017semantic,yu2018generative}. Meanwhile, GANs are notoriously hard to train, and training an image generative model generally requires thousands of images and tens of hours of training time. Actually, for many real-world applications, data acquisition is difficult or expensive. For example, in the artistic domain, it is impossible to hire artists to make thousands of creations. 
Without enough training data, GANs are prone to overfit and collapse. 

To address this issue, researchers begin to focus on effective GAN training with limited samples. Most of them follow the route of few shot generative model adaption that
starting with a model pre-trained on a large dataset of a source domain, and adapting to the target domain with limited data, as shown in Fig.~\ref{fig:setting}.
Wang \etal \cite{wang2018transferring} leverage the fine-tuning strategy to directly model the distribution of target domain.
Some works either impose strong regularization \cite{li2020few,zhang2019consistency} or
modify the network parameters with a slight perturbation
\cite{wang2020minegan,noguchi2019image,robb2020few} to avoid overfitting to the limited target samples.
In addition, some data augmentation methods \cite{tran2021data,zhao2020differentiable,zhao2020image} are proposed to enlarge the amount of the training data so as to improve the robustness of generative models. However, these methods are only suitable for scenarios with more than 100 training images. When the number of training images is reduced to just a few (less than 10), the generative model usually generates images with poor quality and suffers from early collapse.



Recently, as the pioneer work, intuited by the contrastive learning, Ojha \etal \cite{ojha2021few} proposed to preserve the relative similarities and differences between instances in the source domain via an instance distance consistency loss (IDC for short). Given only 10 images, this method can generate more diverse and realistic images for the target domain. Although IDC has made great strides, the generated images still undergo identity degradation and unnatural distortions or textures. The main reason is that IDC can not guarantee the inherent structure of each image, leading to the drift of the samples in the space of target domain (see Sec. \ref{sec:sscl} for a detailed discussion).




In this work, we propose a relaxed spatial structural alignment (RSSA) method to cope with the few shot generative model adaption task. It leverages richer spatial structure priors of images from source domain to address the identity degradation problem of the generative model.
Specifically, we design a cross-domain spatial structural consistency loss, which consists of \textit{self-correlation consistency loss} and \textit{disturbance correlation consistency loss}. The former helps align the self-correlation information of feature maps of the synthesis pairs generated by the source and target generators, so as to constrain the cross-domain consistency of inherent structural information. The latter helps align the 
spatial mutual correlation
between samples adjacent to each other in the latent space, in order to constrain the cross-domain consistency of variation tendency of a specific instance. 

With the help of the cross-domain spatial structural consistency loss, the samples from the target generator maintain original self-correlation and disturbance correlation properties inherited from the source domain during adaption.
However, straightforward alignment may result in the dominant of the attributes from the source domain in the optimization phase,
and slow down the model convergence. Thus, we propose to compress the latent space into a subspace which is close to the target domain. This can relax the above alignment because synthesis pairs generated from the subspace get closer to each other.


To better evaluate the few shot generative model adaption methods, besides the traditional quantitative metric and qualitative visualization, 
we design a structural consistency score (SCS) which measures the structural similarities of synthesis pairs from the source and target domains. Moreover, compared with Inception Score (IS \cite{salimans2016improved}) or Fréchet Inception Distance (FID \cite{heusel2017gans}), SCS can better reflect image identity preservation in few shot adaption.


The main contributions are summarized as follows:
\begin{itemize}[itemsep= 0 pt,topsep = 0 pt, parsep = 0 pt]
	\item We propose RSSA, a relaxed spatial structural alignment method, to transfer rich spatial structural information of the large-scale source domain to the few shot target domain with better identity preservation.
	
	\item We introduce the latent space compression to relax the cross-domain alignment via pulling synthesis pairs generated from the compressed subspace closer to each other, and accelerate the training procedure.

	\item We design a metric to evaluate the quality of synthesis images from the structural perspective, which can serve as an alternative supplement to the current metrics. Qualitatively and quantitatively, our method outperforms existing competitors in a variety of settings. 

\end{itemize}

\vspace{-0.6mm}
\section{Related Work}
\label{sec:related}

\vspace{-0.6mm}
\subsection{Few shot image generation}
\vspace{-1.4mm}
Few-shot image generation aims to generate diversified and high-quality images in a new domain with a small amount of training data. The most straightforward approach is to fine-tune a pre-trained GAN \cite{wang2018transferring,bartunov2018few,liang2020dawson,clouatre2019figr}. However, fine-tuning the entire network weights often leads to poor results. Researchers proposed to modify part of the network weights \cite{mo2020freeze,robb2020few} or batch statistics \cite{noguchi2019image}, and besides leverage different forms of regularization \cite{li2020few,zhang2019consistency} to avoid overfitting. Wang \etal \cite{wang2020minegan} introduced a miner network to steer the sampling of the latent distribution to the target distribution. Data augmentation strategies \cite{tran2021data,zhao2020image,zhao2020differentiable} were introduced to enlarge the amount of the training data to improve the robustness of the generative model. However, most of them fail in the extremely few shot setting (less than 10 images). Recently, Ojha \etal \cite{ojha2021few} proposed to preserve the relative similarities and differences between instances in the source domain via an instance distance consistency loss. Different from work~\cite{ojha2021few}, we explore align the distributions of the source and target domains from the perspective of spatial structural consistency, 
and solve the problems of identity degradation and image distortion during model adaption.

\vspace{-0.6mm}
\subsection{Image to image translation}
\vspace{-1.4mm}
The goal of the image to image translation is to convert an input image from a source domain to a target domain with the intrinsic source content preserved and the extrinsic target style transferred \cite{pang2021image}. Variational autoencoders (VAEs) and GANs are most commonly used and efficient deep generative models in the image-to-image translation tasks \cite{lee2018diverse,ma2018exemplar,isola2017image,lin2020tuigan,fang2020identity,liu2022unsupervised}. However, most methods require a large amount of training data for both source and target domains. Furthermore, generally they are not suitable for the few-shot scenarios. Recent works \cite{liu2019few,wang2020semi,saito2020coco} have begun to address this issue via learning to separate the content and style factors, but require a large amount of labeled data (class or style labels). Different from image to image translation, we explore model-level adaption to the few shot target domain rather than image-level translation. In addition, we do not rely on additional labeled data.



\begin{figure*}[t]
  \centering
   \includegraphics[width=0.985\linewidth]{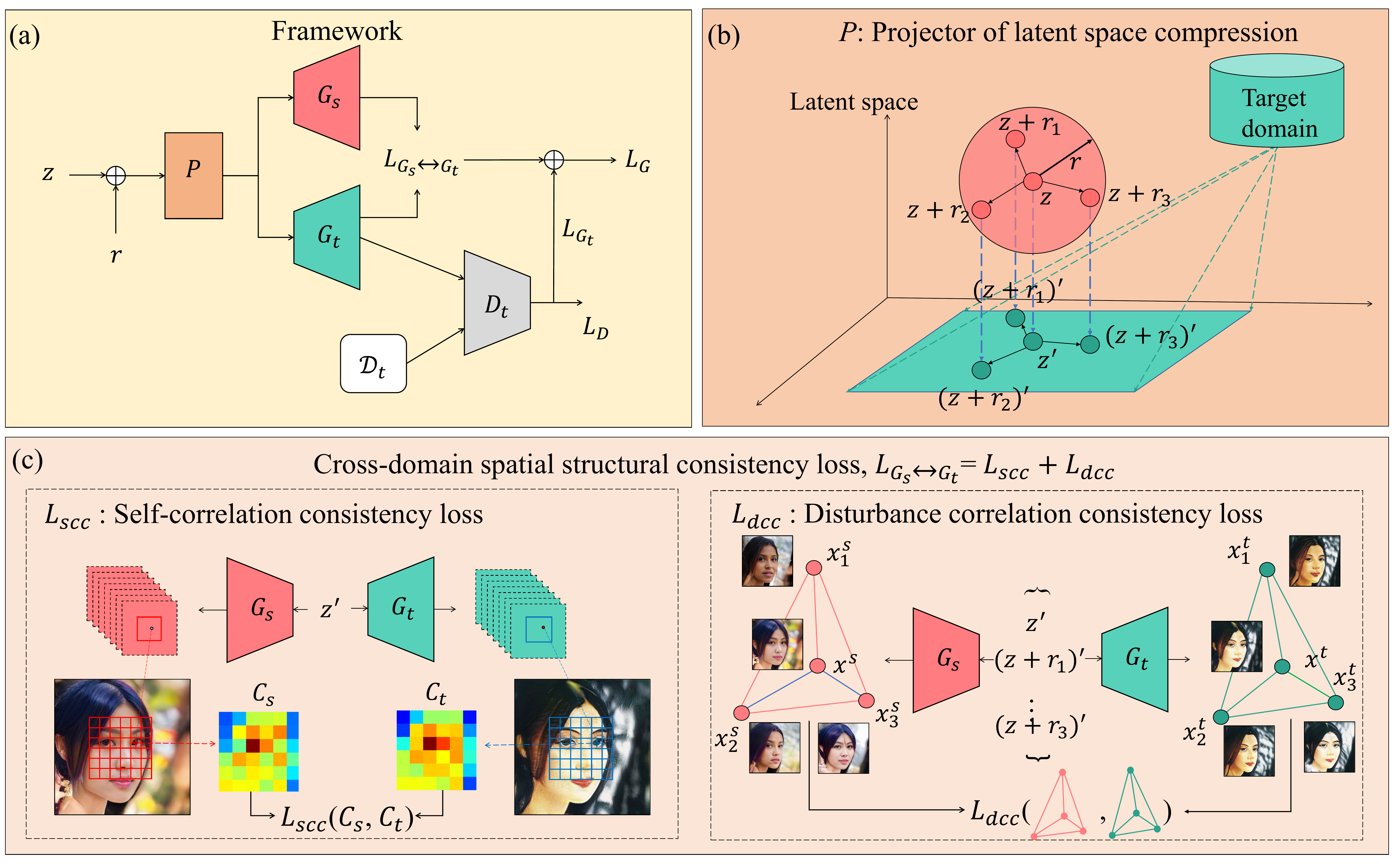}
    \vspace{-3.5mm}   
   \caption{The framework of our method. (a) an overview. 
   $r$ is a small disturbance corresponding to $z$. $P$ is a projector of compressing the latent space as shown in (b). $L_{G_s \leftrightarrow G_t}$ is a cross-domain spatial structural consistency loss, consisting of self-correlation consistency and disturbance correlation consistency loss, as shown in (c).}
   \label{fig:method}
   \vspace{-4.5mm}
\end{figure*}
\vspace{-0.1cm}
\section{Method}
\label{sec:method}
\vspace{-0.15cm}
In this section, we first overview the definition of few shot generative model adaption, and propose the framework of our method. Then, the two key components of RSSA, cross-domain spatial structural consistency loss and latent space compression are interpreted in detail in Sec. \ref{sec:sscl} and \ref{sec:lsc}, the optimization strategy is described in Sec. \ref{sec:optimization}. 
Finally, a novel metric of structural consistency score is proposed to better evaluate few shot generation methods in Sec. \ref{sec:ep}. 
\vspace{-0.7mm}
\subsection{Overview and framework}
\vspace{-1.3mm}
\label{sec:framework}
We have a generator $ G_{s} $ pre-trained on a large-scale dataset in the source domain $ \mathcal{D}_s $. It can be considered as a function that maps a noise vector $z$ sampled from the $d$-dimentional latent space $z\sim p(z)\subset \mathbb{R}^d$ to a generated image $ G_{s}(z) $ in the pixel space. The goal of few shot adaption is to adapt $G_s$ from the source domain to the target domain and obtain $G_{t}$, using a few samples in the target domain. A standard fine-tuning approach is done by initializing $G_{t}$ with $G_s$ and fine-tuning $G_{t}$ on the dataset in the target domain $\mathcal{D}_t$ with an adversarial training procedure. The optimization objectives are as below:
\vspace{-2mm}
\begin{equation}
L_{G}=-{\mathbb{E}}_{z\sim p(z)}[\log(D(G_{t}(z)))]
  \label{eq:L_g}
\end{equation}
\vspace{-5mm}
\begin{equation}
\small
L_{D}={\mathbb{E}}_{x\sim \mathcal{D}_t}[\log(1-D(x)]+{\mathbb{E}}_{z\sim p(z)}[\log(D(G_{t}(z)))],
\label{eq:L_gan}
\small
\end{equation}
where $D$ represents a learnable discriminator.


Most of fine-tuning methods are easy to overfit in the extremely few shot setting, because the discriminator can memorize the few examples and force the
generator to reproduce them. To solve this problem, we consider two strategies. One is preserving the useful structure priors of images from the source domain to restrict the generated images, so as to avoid identity degradation during adaption. The other is compressing the latent space to a subspace where synthesis images of the source and target domains are pulled closer to each other to relax the structural constraint.
Fig.~\ref{fig:method}(a) shows our pipeline. During the model adaption, we conduct relaxed spatial structural alignment. A cross-domain spatial structural consistency loss $L_{G_s \leftrightarrow G_t}$ (Fig.~\ref{fig:method}(c)) and a projector $P$  of latent space compression (Fig.~\ref{fig:method}(b)) are introduced. 

\vspace{-0.9mm}
\subsection{C\hspace{-0.2pt}r\hspace{-0.2pt}o\hspace{-0.2pt}s\hspace{-0.2pt}s\hspace{-0.2pt}-\hspace{-0.2pt}d\hspace{-0.2pt}o\hspace{-0.2pt}m\hspace{-0.2pt}a\hspace{-0.2pt}i\hspace{-0.2pt}n\;s\hspace{-0.2pt}p\hspace{-0.2pt}a\hspace{-0.2pt}t\hspace{-0.2pt}i\hspace{-0.2pt}a\hspace{-0.2pt}l\;s\hspace{-0.2pt}\hspace{-0.2pt}t\hspace{-0.2pt}r\hspace{-0.2pt}u\hspace{-0.2pt}c\hspace{-0.2pt}t\hspace{-0.2pt}u\hspace{-0.2pt}r\hspace{-0.2pt}a\hspace{-0.2pt}l\;c\hspace{-0.2pt}o\hspace{-0.2pt}n\hspace{-0.2pt}s\hspace{-0.2pt}i\hspace{-0.2pt}s\hspace{-0.2pt}t\hspace{-0.2pt}e\hspace{-0.2pt}n\hspace{-0.2pt}c\hspace{-0.2pt}y\;l\hspace{-0.2pt}o\hspace{-0.2pt}s\hspace{-0.2pt}s}
\label{sec:sscl}
\begin{figure}[t]
  \centering
   \includegraphics[width=0.99\linewidth]{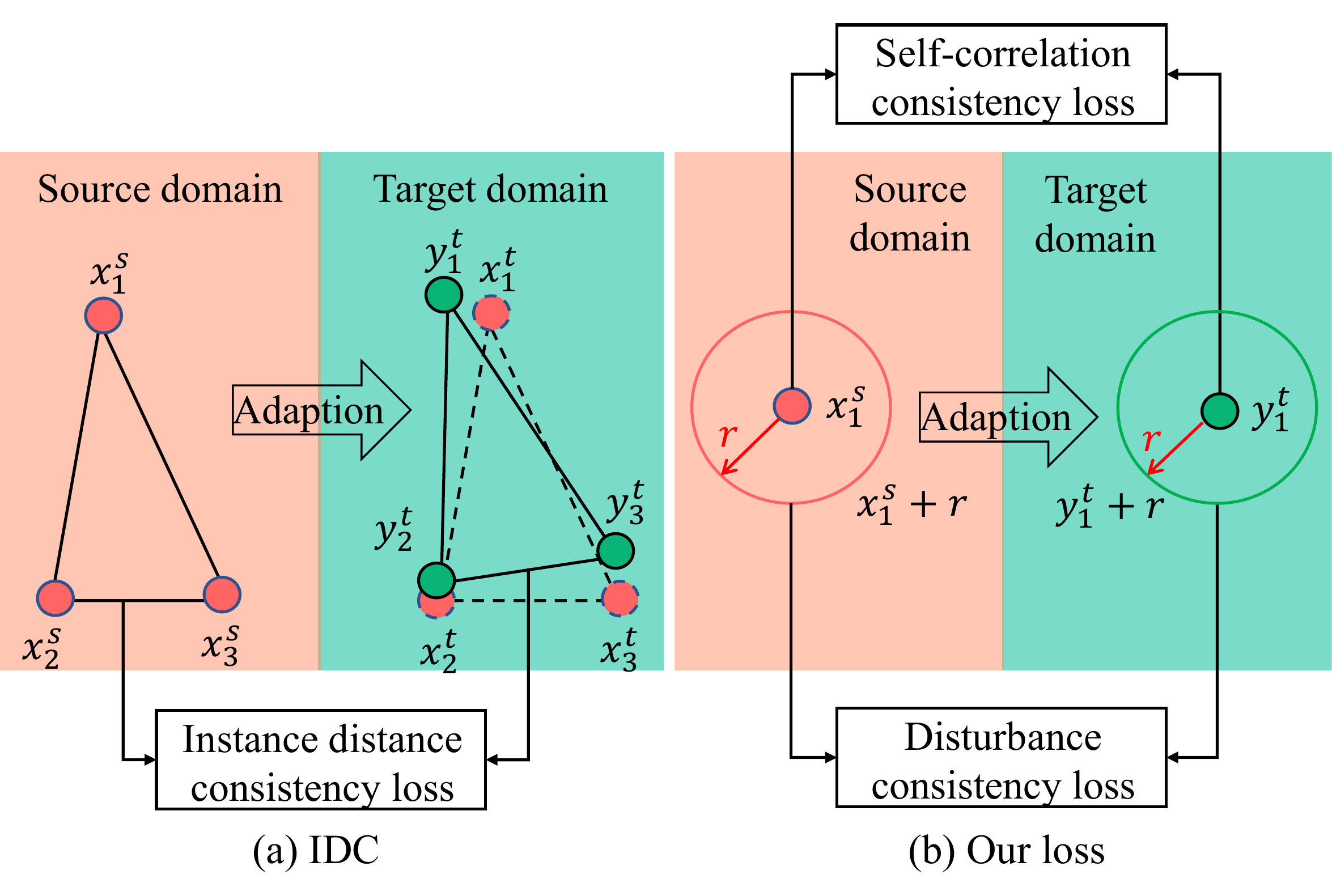}
    \vspace{-2.7mm}
   \caption{Illustration of the spatial structural alignment. (a) instance distance consistency loss from IDC \cite{ojha2021few}. (b) our cross-domain spatial structural consistency loss.}
   \label{fig:motivation}
   \vspace{-5.0mm}
\end{figure}

\vspace{-1mm}
IDC \cite{ojha2021few} preserves the relative distances between instances in the source domain and achieves the state-of-the-art (SOTA) performance in few shot setting. However, the structure of generated images distorts, leading to the problem of identity degradation. Abundant evidences can be found in Fig.~\ref{fig:comp_total}, \ref{fig:comp_10shot} and \ref{fig:comp_5shot}. The main reason is that IDC can not guarantee the inherent structure of each image, leading to the drift of the generated samples in the target domain space. As shown in Fig.~\ref{fig:motivation}(a), the generated images of the target domain (the green points) maintain the correct instance distances, but deviate from their correct positions (the red points). To avoid this drift, we propose a cross-domain spatial structural consistency loss, which preserves the inherent spatial structure and the variation tendency of images from the source domain as shown in Fig.~\ref{fig:motivation}(b). Specifically, we design a self-correlation consistency loss to constrain the inherent structure of the images, and a disturbance correlation consistency loss to constrain the variation tendency under a certain disturbance of the images.



\noindent{\bf Self-correlation consistency loss.} 
We are inspired by some relevancy mining methods\cite{liu2022unsupervised,li2022Long,li2016distributed} 
and adopt the self-correlation matrices of feature maps at each each convolutional layer to formulate the inherent structure information of the image. Each pair of self-correlation matrices from the same layers of $G_s$ and $G_{t}$ are constrained with the $\text{smooth-}{\ell}_{1}$ loss \cite{girshick2015fast}, so as to ensure that the images in source and target domains have similar inherent structure. Define $f^l\in \mathbb{R}^{c\times w \times h}$ as the feature maps at the $l^{th}$ layer. $f^l(x,y)$ is a $c$ dimensional vector. Each entry of the self-correlation matrix $C^l_{x,y}\in \mathbb{R}^{w\times h}$ of the position $(x,y)$ at the $l^{th}$ layer can be calculated as below:
\begin{equation}
  C^l_{x,y}(i,j)=\text{cos}(f^l(x,y),f^l(i,j)),  \label{eq:cos_sim}
\end{equation}
\vspace{-1.0mm}
where cos($\cdot$) denotes the cosine similarity function, $(i,j)$ is the corresponding position in $f^l$.
The spatial self-correlation consistency loss between $G_s$ and $G_{t}$ can be calculated as
\begin{equation}
\small
L_{\text{scc}}(G_{t},G_{s})={\mathbb{E}}_{z_i\sim p(z)}\sum_{l}\sum_{x,y}\text{smooth-}{\ell}_{1}(C_{x,y}^{t,l}, C_{x,y}^{s,l}),
  \label{eq:L_scc}
\end{equation}
\vspace{-1.0mm}
where $C_{x,y}^{t,l}$ and $C_{x,y}^{s,l}$ indicate the self-correlation matrices of the position $(x,y)$ at the $l^{th}$ layer for $G_{t}$ and $G_{s}$.

Note that computation of the self-correlation matrices is an $O((w\cdot h)^2)$ operation. For feature maps with high resolution, we first aggregate the adjacent feature vectors by adopting average pooling and break the whole feature map into patches to compute local self-correlation matrices.


\noindent{\bf Disturbance correlation consistency loss.} The latent space of a generative model is continuous rather than discrete, hence we propose to model the variation tendency under certain disturbances of the images(essentially the gradient information around each
instance). Specifically, we take an input noise vector as an anchor point, and then sample a batch of vectors from a small neighborhood of this anchor point. The spatial similarities between these samples are calculated and transferred from the source domain to the target domain. 
%



For an input noise $z_i$, define a neighborhood with radius $r$, $U(z_i,r)=\{z||z-z_i|<r\}$. We sample $N$ noise vectors from $U(z_i,r)$ and form a batch of $N+1$ vectors $\left \{ z_{n} \right \}_{1}^{N+1}$ to represent the neighborhood. Define $D^l_{jk}$ as the pixel-wise spatial mutual correlation for the $l^{th}$ layer feature map $f^l$ between any two samples $z_j$ and $z_k$ from $\left \{ z_{n} \right \}_{1}^{N+1}$. 
$D^l_{jk}$ at position $(x,y)$ is denoted by the softmax of similarities between feature vector at $(x,y)$ in $f^l_j$ and a small corresponding region {\small{$Q=\{(m,n)|x-\frac{\delta}{2}<m<x+\frac{\delta}{2},y-\frac{\delta}{2}<n<y+\frac{\delta}{2}\}$}}, where $\delta$ is the width of a slide window, in $f^l_k$ as below:
\begin{equation}
\small
  D^l_{jk}(x,y)=\text{Softmax}(\{\text{cos}(f^l_j(x,y),f^l_k(m,n))\}_{(m,n)\in Q}),
  \label{eq:Disturbance}
\small
\end{equation}
where cos($\cdot$) denotes the cosine similarity function.

On basis of the computed pixel-wise correlation distribution, we impose the disturbance correlation consistency constraint by minimizing the L1 distance:
\begin{equation}
\footnotesize
L_{\text{dcc}}(G_{t},G_{s})={\mathbb{E}}_{z_i\sim p(z)}\sum_{l,j,k,x,y}\left \| D^{t,l}_{jk}(x,y)-D^{s,l}_{jk}(x,y) \right \|_1,
\footnotesize
  \label{eq:L_dcc}
\end{equation}
The spatial structural consistency loss $L_{G_s \leftrightarrow G_t}$ is composed of the self-correlation consistency loss $L_{\text{scc}}$ and disturbance correlation consistency loss $L_{\text{dcc}}$. It is calculated as below:
\begin{equation}
L_{G_s \leftrightarrow G_t} =\alpha L_{\text{scc}} + \beta L_{\text{dcc}}.
  \label{eq:L_gst}
\end{equation}
where $\alpha$, $\beta$ are ratio parameters.




\vspace{-0.6mm}
\subsection{Latent space compression}
\label{sec:lsc}
\vspace{-1.2mm}
The spatial structural consistency loss helps align the generated images of the source and target domains. However, straightforward alignment may cause the dominant of the attributes from the source domain, and thus slow down model adaption. Therefore, we propose to compress the latent space to a subspace close to the target domain to relax the cross-domain alignment. Specifically, we first invert the few samples from target domain $\{x^t_i\}_{i=1}^n$ to the $W^+$ space of $G_s$ by utilizing Image2StyleGAN \cite{abdal2019image2stylegan}. Given $n$ target samples, an inverted latent code set at $l^{th}$ layer is denoted as $\{w^l_i\}_{i=1}^n$. Define a $n$ column matrix $A^l$ which is composed of $\{w^l_i\}_{i=1}^n$ by $A^l_{*i}=w^l_i$, and hence we obtain a subspace $\mathcal{X}^l$ of the $l^{th}$ latent space, where $\mathcal{X}^l$ is equivalent to the column space of $A^l$. Given an input noise $z_j$, the corresponding latent code in the $l^{th}$ layer is $w_j^l$, the corresponding modulation coefficient is $\alpha^{l}$, we modulate $w_j^l$ and project it onto the $l^{th}$ sub-plane $\mathcal{X}^l$ via least square method:
\begin{equation}
\begin{aligned}
    \overline{w_j^l}&=A^l({A^l}^{\top}A^l)^{-1}{A^l}^{\top}w_j^l \\
    \hat{w_j^l}&=\alpha^{l}\overline{w_j^l}\frac{\left \| w_j^l \right \| }{\left \| \overline{w_j^l}  \right \|} +(1-\alpha^{l})w_j^l,
\end{aligned}
  \label{eq:latent space compression}
\end{equation}
where $\hat{w_j^l}$ is the projected code at $l^{th}$ for $z_j$.

\begin{figure}[t]
  \centering
   \includegraphics[width=0.99\linewidth]{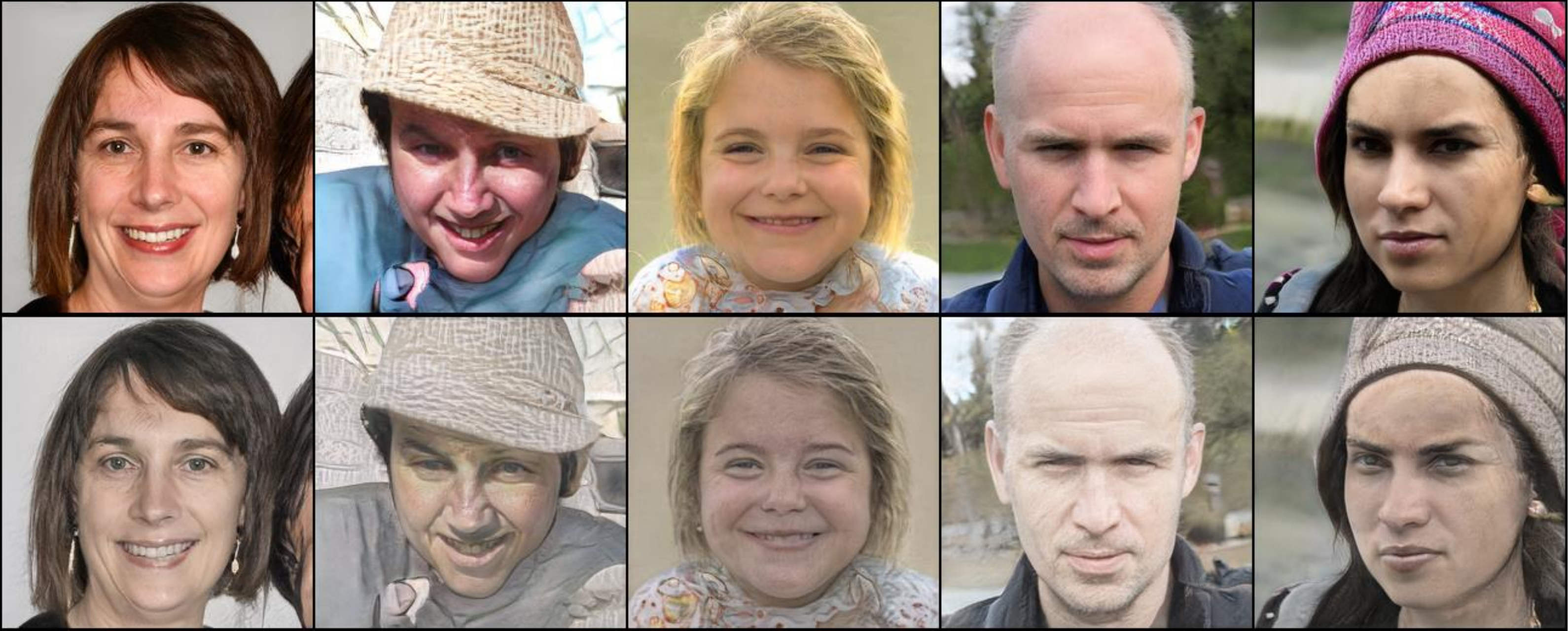}
   
   \caption{Generated images of $G_s$ with input latent codes sampled from different spaces. Top: original latent space. Bottom: compressed subspace. Setting: Flickr-Faces $\rightarrow$ Sketches.}
   \label{fig:projection}
   \vspace{-3mm}
\end{figure}

In this way, we compress the original latent space into a narrow subspace close to the target domain. Images generated by $G_s$ with the latent codes sampled from the compressed subspace imply some characteristics of the target domain. As shown in Fig.~\ref{fig:projection}, generated images (bottom row) shows some characteristics of sketch on the texture and color. By sampling latent codes from the compressed subspace before the alignment, we are capable to stabilize and accelerate the whole training procedure. Note that latent codes of different layers modulate the output images at distinct semantic levels in StyleGAN \cite{karras2019style} architecture. We set large $\alpha_i$ at the top layers and small ones at the bottom layers to alter the generated images' attributes at high semantic levels while maintaining the original spatial structure to a great extent.  

\subsection{Optimization}
\label{sec:optimization}

We follow an adversarial optimization procedure. The objective of $G_t$ is a combination of $L_G$ (Eq.\ref{eq:L_g}) and $L_{G_s \leftrightarrow G_t}$ (Eq.\ref{eq:L_gst}). We simply set the hyper-parameters $\alpha$ and $\beta$ as 1 for all experiments. The objective of $D$ is the same with \cite{ojha2021few} by utilizing a combination of image-wise and patch-wise discriminant loss. 
Standard path regularization loss to $G_t$ and gradient penalization loss to $D$ are also adopted at every several iterations. 

\subsection{Evaluation metric}
\label{sec:ep}

In addition to the intuitively visual evaluation, IS \cite{salimans2016improved} and FID \cite{heusel2017gans} are the most widely used quantitative evaluation metrics for image generative models. However, both of them map the generated images to the feature space by an Inception network, which can not quantify the quality of the spatial structure of the generated images. Meanwhile, in few shot setting, many fine-tuning based methods are inclined to simply synthesis images similar with the training samples given arbitrary input noises. Yet, this may obtain high IS in some cases which is counter-intuitive, see Table \ref{tab:quan_eval}.
Furthermore, computing FID requires a large number of realistic images of the target domain, which is impractical in the few shot setting. Therefore, we adopt IS as a general evaluation metric and propose a novel spatial structure evaluation metric, termed structural consistency score, to cover the shortage of IS.


\noindent\textbf{Structural Consistency Score (SCS).}  
For a image pair $\left \langle x^s, x^t  \right \rangle $ generated by $\left \langle G_s, G_t  \right \rangle $ with the same input $z_i$, we claim that $x^t$ preserves structural consistency when it can be easily recognized as a derivative sample from $x^s$. Inspired by \cite{yi2020unpaired}, we extract a meaningful edge map of one image to represent its structural information by HED\cite{xie2015holistically}. The SCS of a generated image of the target domain $x^t$ is computed with the dice similarity coefficient \cite{dice1945measures} between the edge maps of $x^t$ and $x^s$. The formalization is as below:
\begin{equation}
\begin{aligned}
    \text{SCS}(x^t)& = \frac{2|H(x^t)\cap H(x^s)|}{|H(x^t)|+|H(x^s)|},
\end{aligned}
  \label{eq:L_HED_DSC}
\end{equation}
where $H(\cdot)$ denotes HED function. $|H(x^t)\cap H(x^s)|$ is calculated by the pixel-wise inner product of $H(x^t)$ and $H(x^t)$. $|H(x^{t})|$ and $|H(x^s)|$ is calculated by the sum of squares of matrix elements. Then, the SCS of the target GAN $G_t$ can be quantified into the expectation as below:
\begin{equation}
\begin{aligned}
    \text{SCS}(G_t)
    =\mathbb{E}_{z_i\sim p(z)}[\frac{2|H(G_t(z_i))\cap H(G_s(z_i))|}{|H(G_t(z_i))|+|H(G_s(z_i))|}].
\end{aligned}
  \label{eq:structrual_consistency}
\end{equation}
Higher SCS means better spatial structural consistency between $G_t$ and $G_s$. It is remarkable that the SCS of an overfitted model will be very low, because it can not generate images with structures similar to the source domain images.


\section{Experiments}
In this section, we demonstrate the effectiveness of RSSA in few shot setting. Qualitative and quantitative comparisons between our method and several baselines, TGAN \cite{wang2018transferring}, FreezeD \cite{mo2020freeze}, MineGAN \cite{wang2020minegan}, IDC \cite{ojha2021few}. As the SOTA method, IDC \cite{ojha2021few} is our primary comparison method in most experiments. 

We adopt the StyleGANv2 \cite{karras2020analyzing} pre-trained on three different datasets: (i) Flickr-Faces-HQ (FFHQ) \cite{karras2019style}, (ii) LSUN Churches \cite{yu2015lsun}, (iii) LSUN Cars \cite{yu2015lsun}. We adapt the source GANs to various target domains including: (i) face sketches \cite{wang2008face}, (ii) face paintings by Van Gogh \cite{yaniv2019face}, (iii) face paintings by Moise Kisling \cite{yaniv2019face}, (iv) haunted houses \cite{ojha2021few}, (v) village painting by Van Gogh \cite{ojha2021few}, (vi) wrecked/abandoned cars \cite{ojha2021few}. Model adaptions are done in 10-shot, 5-shot and 1-shot settings.

\begin{figure*}[t]
  \centering
   \includegraphics[width=0.99\linewidth]{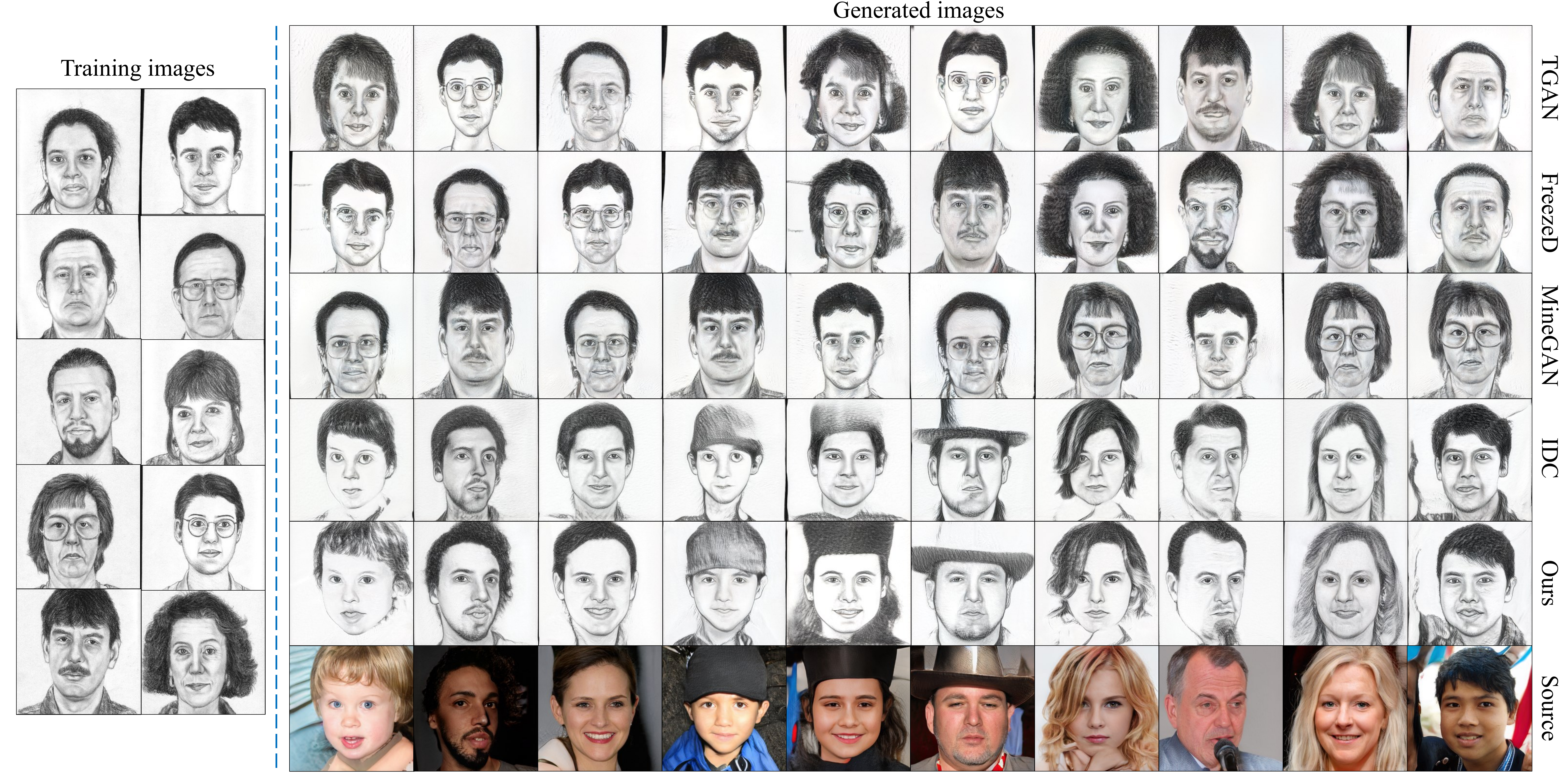}

   \vspace{-1mm}
   \caption{Comparison results with different methods on Flickr-Faces $\rightarrow$ Sketches (10-shot).}
   \label{fig:comp_total}
   \vspace{-1mm}
\end{figure*}

\begin{figure*}[t]
  \centering
   \includegraphics[width=0.99\linewidth]{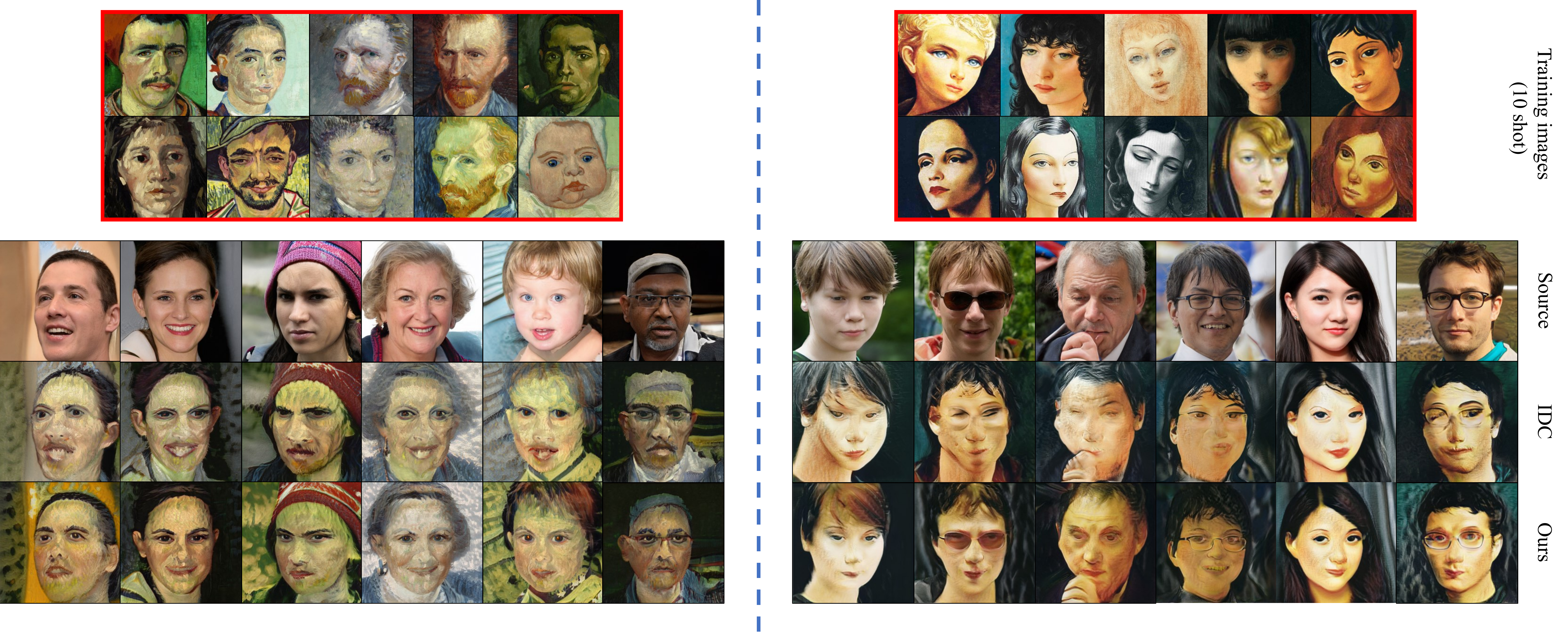}

   \vspace{-3mm}
   \caption{Comparison results between IDC and our method on Flickr-Faces $\rightarrow$ Vincent van Gogh (left), Moise Kisling (right) (10-shot).}
   \label{fig:comp_10shot}
   \vspace{-2mm}
\end{figure*}

\begin{figure*}[t]
  \centering
   \includegraphics[width=0.99\linewidth]{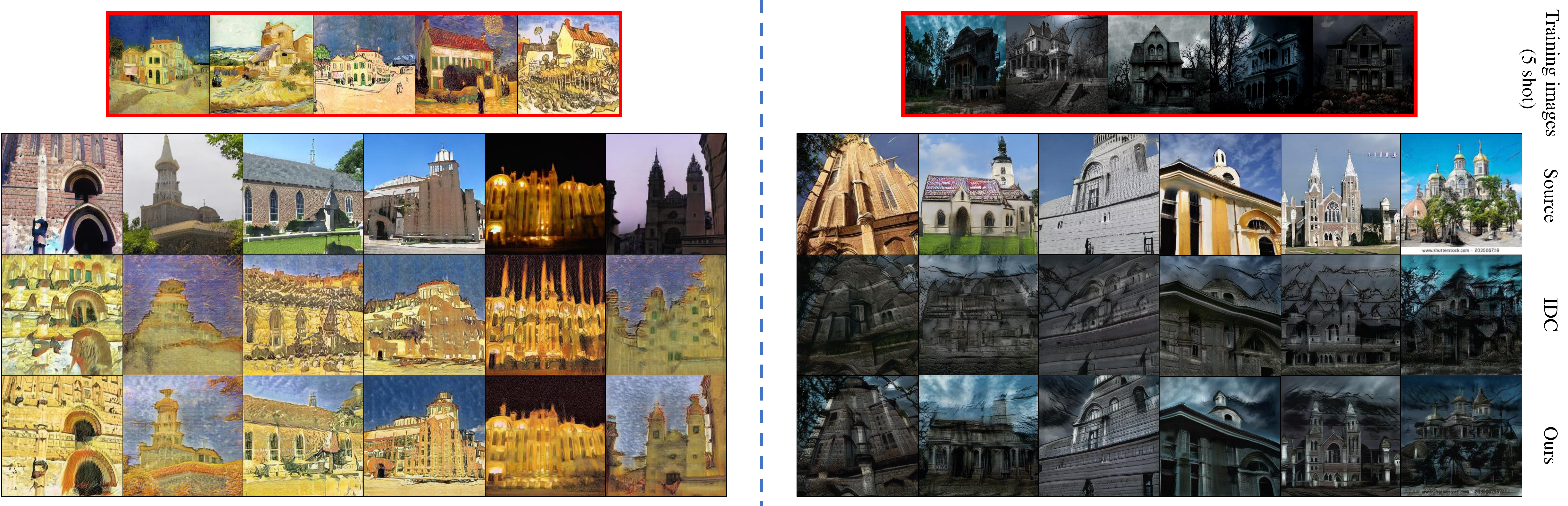}
   \vspace{-2mm}
   \caption{Comparison results between IDC and our method on Churches $\rightarrow$ Van Gogh Village (left), Haunted Houses (right) (5-shot).}
   \label{fig:comp_5shot}
   \vspace{-2mm}
\end{figure*}

\begin{table*}[t]
  \centering
  \renewcommand\arraystretch{1.2}
\resizebox{\textwidth}{!}{
\begin{tabular}{llllllllllll}
\hline
\multicolumn{1}{l}{\multirow{2}{*}{Metric}} & \multicolumn{1}{l}{\multirow{2}{*}{Method}} & \multicolumn{2}{c}{f$\rightarrow$S} & \multicolumn{2}{c}{f$\rightarrow$V} & \multicolumn{2}{c}{f$\rightarrow$M} & \multicolumn{2}{c}{c$\rightarrow$V} & \multicolumn{2}{c}{c$\rightarrow$H}  \\ \cline{3-12} 
  \multicolumn{1}{c}{}   &   \multicolumn{1}{c}{}     & {10-shot}            & {5-shot}           & {10-shot}            & {5-shot}           & {10-shot}     & {5-shot}    & {10-shot}            & {5-shot}           & {10-shot}            & {5-shot}          \\ \hline
\multirow{5}{*}{IS}     & TGAN                    & 2.00$^{\pm0.04}$       & 2.09$^{\pm0.05}$             & \textbf{2.69$^{\pm0.18}$ }    & 1.65$^{\pm0.07}$         & 1.88$^{\pm0.08}$        & 1.98$^{\pm0.04}$      & 2.82$^{\pm0.09}$       & 2.28$^{\pm0.05}$       & 5.16$^{\pm0.19}$       & 4.47$^{\pm0.14}$        \\
                        & FreezeD                 & 1.98$^{\pm0.05}$       & 2.05$^{\pm0.06}$             & 2.46$^{\pm0.06}$     & \textbf{1.77$^{\pm0.03}$ }        & 1.91$^{\pm0.06}$        & 1.85$^{\pm0.04}$       & 2.78$^{\pm0.07}$       & 2.32$^{\pm0.09}$       & 5.29$^{\pm0.15}$       & 4.60$^{\pm0.17}$        \\
                        & MineGAN                 & 1.92$^{\pm0.03}$       & 2.12$^{\pm0.07}$             & 2.52$^{\pm0.09}$     & 1.71$^{\pm0.04}$         & 1.84$^{\pm0.03}$        & 1.87$^{\pm0.08}$      & 2.51$^{\pm0.05}$       & 2.18$^{\pm0.08}$       & 5.22$^{\pm0.11}$       & 4.55$^{\pm0.18}$        \\
                        & IDC                     &1.95$^{\pm0.02}$           &2.15$^{\pm0.04}$ & 1.55$^{\pm0.04}$      & 1.52$^{\pm0.02}$  & 1.97$^{\pm0.03}$    & 
                        1.76$^{\pm0.06}$
                        & 2.86$^{\pm0.11}$ & 2.78$^{\pm0.06}$  &  5.44$^{\pm0.21}$   & 5.15$^{\pm0.13}$                 \\
                        & Ours                    & \textbf{2.10$^{\pm0.03}$ }  & \textbf{2.41$^{\pm0.03}$ } & 1.77$^{\pm0.06}$  &1.61$^{\pm0.04}$ & \textbf{2.17$^{\pm0.07}$ } &  \textbf{2.09$^{\pm0.05}$ } &\textbf{3.54$^{\pm0.10}$} & \textbf{3.62$^{\pm0.13}$}                 &\textbf{6.26$^{\pm0.18}$} &\textbf{5.70$^{\pm0.20}$}  \\ \hline
\multirow{5}{*}{SCS}    & TGAN                    & 0.287              & 0.285            & 0.380              & 0.375            & 0.344       & 0.350         & 0.355              & 0.343            & 0.211              & 0.218            \\
                        & FreezeD                 & 0.289              & 0.288            & 0.384              & 0.372            & 0.347       & 0.346         & 0.353              & 0.349            & 0.216              & 0.212            \\
                        & MineGAN                 & 0.294              & 0.290            & 0.340              & 0.332            & 0.339       & 0.341         & 0.375              & 0.350            & 0.211              & 0.214            \\
                        & IDC                     &0.437& 0.422                 & 0.594             &0.568            & 0.524      &   0.494     &0.551                  & 0.535            & 0.490               & 0.424          \\
                        & Ours                    &\textbf{0.511}&                   \textbf{0.507} & \textbf{0.702}          &     \textbf{0.685}          & \textbf{0.644}        &  
                        \textbf{0.618}
                        &\textbf{0.702} & \textbf{0.681}                & \textbf{0.573}             & \textbf{0.582 }        \\ \hline

\end{tabular}
}

  \centering
  \vspace{-1mm}
  \caption{Quantitative evaluation of methods by IS and SCS. f and c represent faces and churches source domains. S,V,M,H represent four target domains: sketches, Van Gogh's paintings, Moise Kisling's paintings and haunted houses. Best results are \textbf{bold}.}
  \label{tab:quan_eval}%
   \vspace{-3mm}
\end{table*}

\subsection{Performance evaluation}\label{subsec:evalutaion}
\noindent\textbf{Qualitative comparison.}  
\begin{figure}[t]
  \centering
   \includegraphics[width=0.99\linewidth]{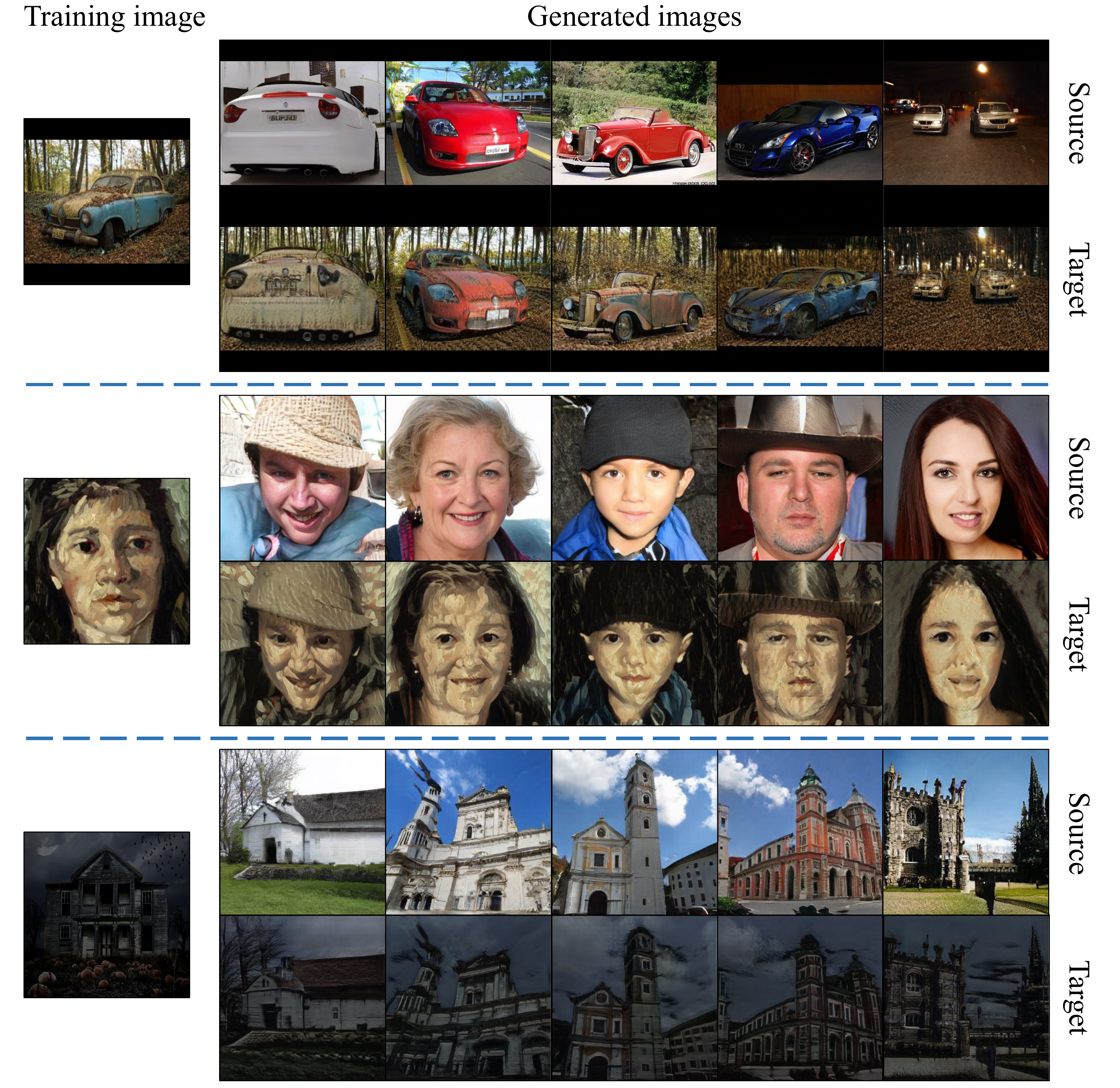}
   \vspace{-2mm}
   \caption{Results of our method on Cars $\rightarrow$ Wrecked Cars, Faces $\rightarrow$ Vincent van Gogh, and Churches $\rightarrow$ Haunted houses (1-shot).
   }
   \label{fig:1shot}
   \vspace{-6mm}
\end{figure}
Fig.~\ref{fig:comp_total} shows results on FFHQ $\rightarrow$ sketches using different adaption methods. We can observe that TGAN overfits strongly to the samples of the target domain. Compared with TGAN, FreezeD and MineGAN do not improve the results. This indicates that although they could play a positive role in a small dataset with more than 100 training samples, they would be ineffective in extremely few shot setting (less than 10). By contrast, IDC improves the correspondence between the source domain and the target domain, and shows similar visual patterns between the synthesis pairs. Further, as for images synthesised by RSSA, one can easily recognize the corresponding source domain images with only few glances. This is because our method acquires visual attributes from the target domain and meanwhile greatly preserves the spatial structural information of images from the source domain.

In order to comprehensively compare IDC with our method, we extend comparison experiments to multiple target domains with different few shot setting as shown in Fig.~\ref{fig:comp_10shot} (10-shot) and Fig.~\ref{fig:comp_5shot} (5-shot). We can observe the distorted attributes of human faces (Fig.~\ref{fig:comp_10shot}) and texture degradation of churches (Fig.~\ref{fig:comp_5shot}) in IDC's results, but there are almost no similar phenomenons in our results. In addition, we take a bold stab at 1-shot scenarios as shown in Fig.~\ref{fig:1shot}, and obtain some decent results. Good visual results are mainly attributed to the cooperation of latent space compression and spatial structure alignment, the former helps acquire attributes from the target domain faster, the latter helps preserve the structural knowledge from the source domain.

\begin{figure}[t]
  \centering
   \includegraphics[width=1\linewidth]{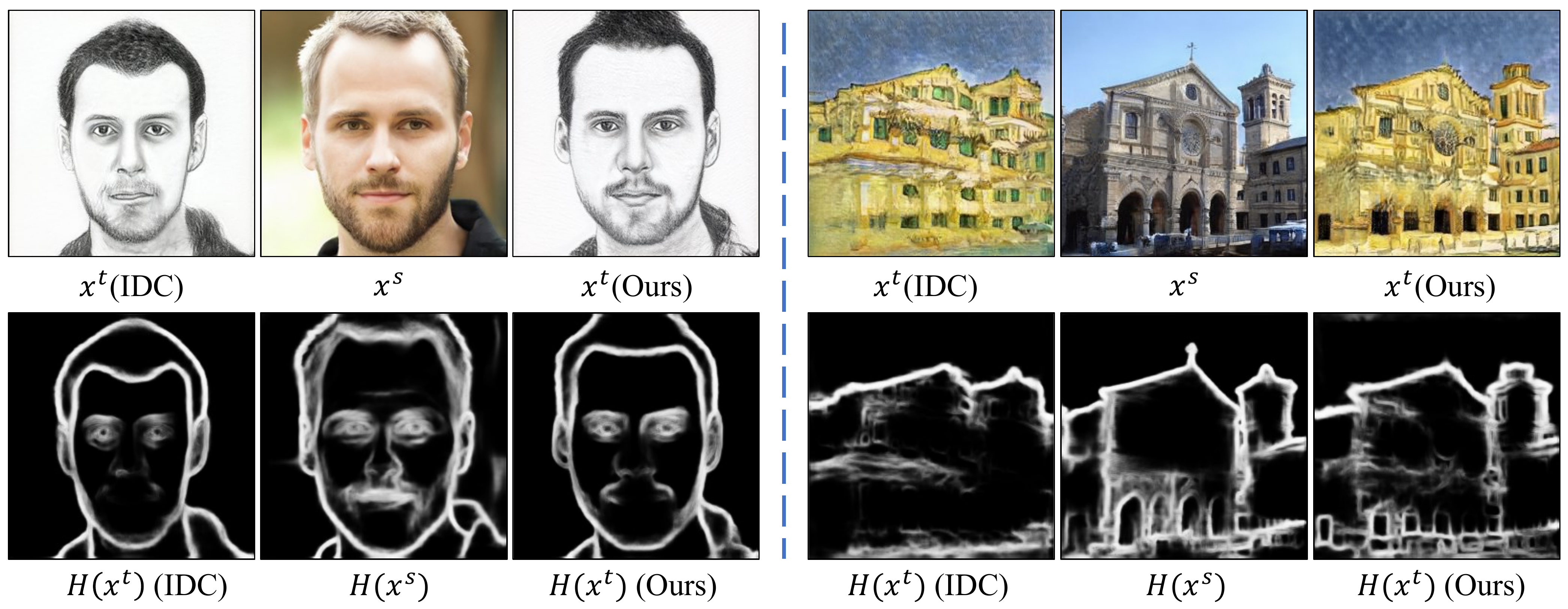}
   \vspace{-5mm}
   \caption{Comparison of edge maps between IDC and our method. Top row shows generated images, bottom raw shows corresponding edge maps. Left: IDC. Middle: Source. Right: Ours}
   \vspace{-4mm}
   \label{fig:HED_eval}
\end{figure}

\noindent\textbf{Quantitative comparison.} To quantify the quality and diversity of the synthesis images, we evaluate all methods with IS and SCS. All quantitative experiments are conducted in 10-shot and 5-shot settings. For IS, we calculate means and variances over 10 runs on 10000 randomly sampled images. As shown in Table \ref{tab:quan_eval}, our method achieves best scores in most cases due to the high diversity and quality of synthesis images. However, TGAN, FreezeD and MineGAN outperforms IDC and RSSA on face$\rightarrow$Van Gogh's paintings. The reason is that they simply overfit to the few training samples, while images generated by IDC and RSSA tend to remix the textures and colors of the paintings. 
This indicates that IS sometimes fails to handle the overfitting problem in few shot setting. For SCS, we randomly sample 500 noise vectors as inputs of $G_s$ and $G_t$, then form the synthesis pairs and calculate their mean score. As shown in Table \ref{tab:quan_eval}, TGAN, FreezeD and MineGAN overfit to the training samples and obtain lower results for all settings. IDC performs much better by preserving the distances of instances. RSSA consistently surpasses all the comparison methods by a large margin due to the better spatial structure preservation. To visually understand SCS, Fig.~\ref{fig:HED_eval} shows two groups of edge extraction examples of on FFHQ $\rightarrow$ sketches and Churches $\rightarrow$ Van Gogh Village. Obviously, our generated images retain more accurate edge information than those from IDC, thus obtain higher SCS.

\subsection{Ablation study}\label{subsec:ablation}
\begin{figure}[h]
  \centering
   \includegraphics[width=1\linewidth]{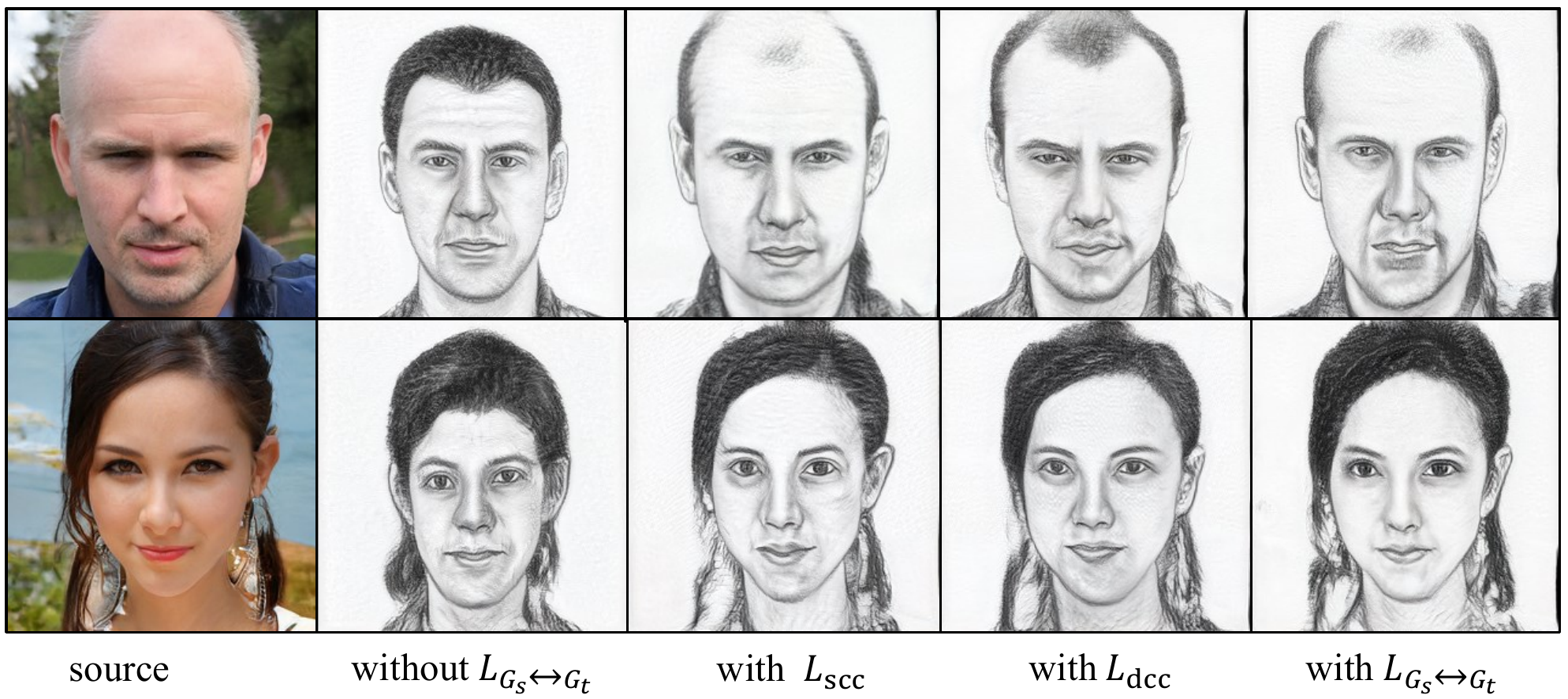}
   \caption{Qualitative ablation of the loss function on FFHQ $\rightarrow$ sketches in 10-shot setting.}
   \vspace{-3mm}
   \label{fig:ablation_visual}
\end{figure}

\noindent\textbf{Effect of the spatial structural consistency loss.} 
We conduct ablation experiments to verify the effectiveness of the two components of our proposed spatial structural consistency loss. As shown in Fig.~\ref{fig:ablation_visual}, $L_{scc}$ and $L_{dcc}$ greatly improve the visual quality of synthesis images separately and get the best visual results in cooperation. Consistently, quantitative conclusion can be drawn in Table~\ref{tab:albation}. The reason is that they introduce inherent and neighbouring structural constraint separately and share good compatibility.


\begin{figure}[t]
  \centering
   \includegraphics[width=1\linewidth]{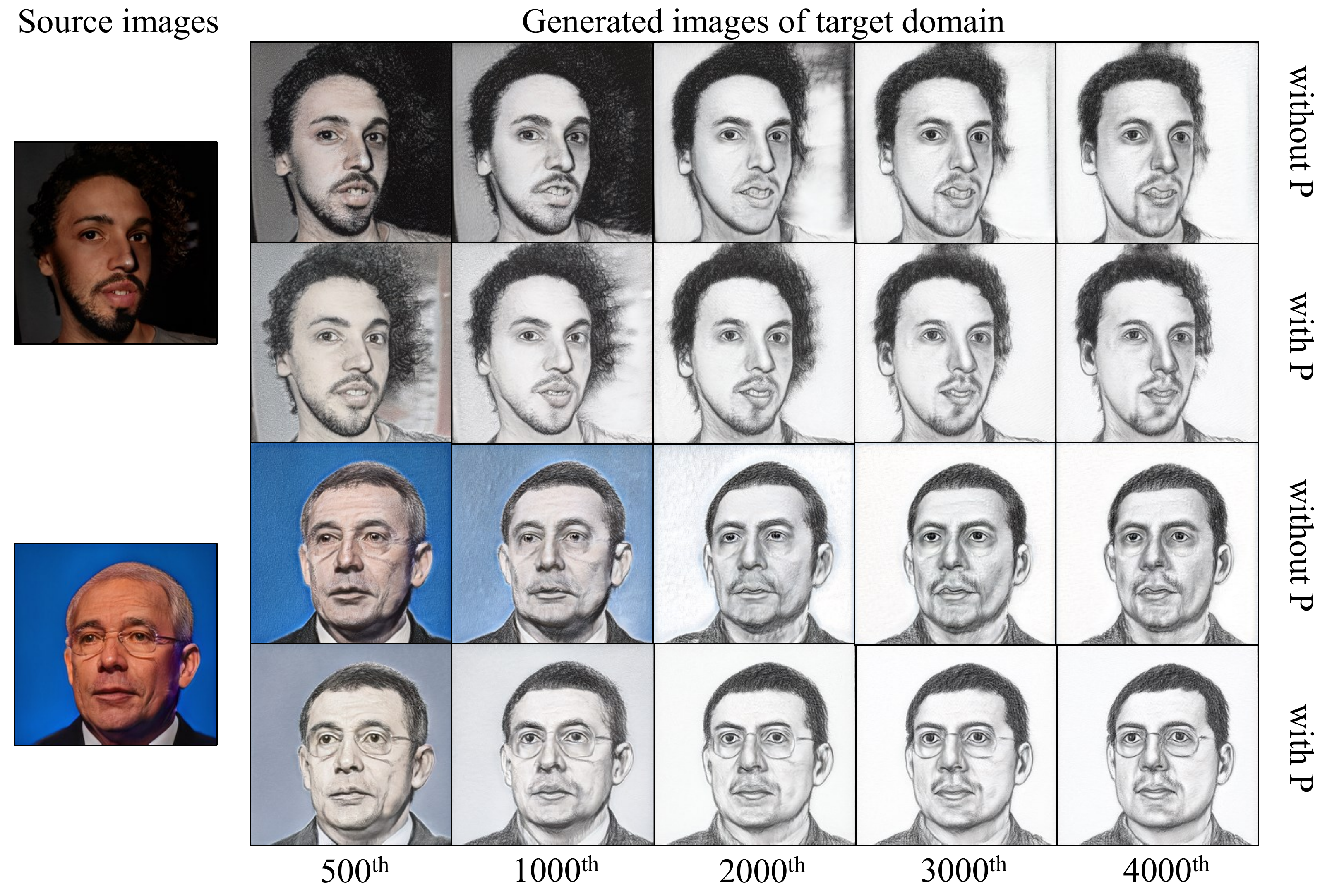}
   \vspace{-5mm}
   \caption{Ablation of the latent space compression, which shows the output images from the same input noises at different training iterations. $P$ denotes the projector of compression.}
   \vspace{-4.5mm}
   \label{fig:ablation_proj}
\end{figure}

\begin{table}[h]
\small
\begin{tabular}{lccccc}
\hline
\multirow{2}{*}{Method} & \multicolumn{2}{c}{IS}                & \multicolumn{2}{c}{SCS}               \\ \cline{2-5} 
                        & f$\rightarrow$V & c$\rightarrow$H & f$\rightarrow$V & c$\rightarrow$H \\ \hline
without $L_{G_s \leftrightarrow G_t}$          & 1.40$^{\pm0.03}$               & 5.38$^{\pm0.09}$             & 0.449              & 0.369              \\
with $L_{\text{scc}}$                & 1.69$^{\pm0.05}$              & 6.18$^{\pm0.17}$             & 0.684               & 0.544             \\
with $L_{\text{dcc}}$                &
1.64$^{\pm0.05}$     & 6.13$^{\pm0.14}$    & 0.671                 & 0.548             \\
with $L_{G_s \leftrightarrow G_t}$               & \textbf{1.77$^{\pm0.06}$}            & \textbf{6.26$^{\pm0.18}$}            & \textbf{0.702}                & \textbf{0.573}             \\ \hline
\end{tabular}
\small
  \caption{Quantitative ablation of the loss function in 10-shot setting. f and c represent faces and churches source domains, V and H represent Van Gogh's paintings and haunted houses target domains.}
   \vspace{-3mm}
  \label{tab:albation}
\end{table}
\noindent\textbf{Effect of the latent space compression.} We conduct the model adaption on FFHQ $\rightarrow$ sketches in 10-shot setting. As shown in Fig.~\ref{fig:ablation_proj}, we observe that some target-domain irrelevant attributes (e.g. background and color) degrade faster when adopting latent space compression. This demonstrates that (1) the latent space compression copes with the dominant of attributes from source domain well so as to relax the spatial structural alignment; (2) it helps speed up the adaption procedure of $G_t$. 

\section{Conclusion and Limitations}

In this paper, we propose a novel few shot generative model adaption method, relaxed spatial structural alignment (RSSA). By aligning the generative distributions of the source and target domains via a cross-domain spatial structural consistency loss, the inherent structure information and spatial variation tendency of images from the source domain can be well preserved and transferred to the target domain. The original latent space is compressed to a narrow subspace close to the target domain, which relaxes the cross-domain alignment and accelerates the convergence rate of the target domain generator. In addition, we design a novel metric, SCS, to assess the structural quality of generated images. It may serve as an alternative supplement to the current metrics in the few shot generation scenarios.

Although our method can deal with the settings of extremely few shot training samples well and generate compelling visual results, it also has some limitations. The spatial structural consistency loss is not friendly to some abstract domains, such as paintings from Amedeo Modigliani, who is known for portraits in a modern style characterized by a surreal elongation of faces. Nevertheless, we believe that more data-efficient generative models will be proposed in the near future. The application of these models will in turn facilitate a wide range of downstream tasks such as few shot image classification.

\noindent\textbf{Acknowledgement.} This work was supported in part by the National Key R\&D Program of China under Grant 2018AAA0102000, and in part by the National Natural Science Foundation of China: U21B2038, U19B2038, 61931008, 61732007, and CAAI-Huawei MindSpore Open Fund, Youth Innovation Promotion Association of CAS under Grant 2020108, CCF-Baidu Open Fund.

{\small
\bibliographystyle{ieee_fullname}
\bibliography{egbib}

\begin{thebibliography}{10}\itemsep=-1pt

\bibitem{abdal2019image2stylegan}
Rameen Abdal, Yipeng Qin, and Peter Wonka.
\newblock Image2stylegan: How to embed images into the stylegan latent space?
\newblock In {\em ICCV}, pages 4432--4441, 2019.

\bibitem{bartunov2018few}
Sergey Bartunov and Dmitry Vetrov.
\newblock Few-shot generative modelling with generative matching networks.
\newblock In {\em AISTATS}, pages 670--678. PMLR, 2018.

\bibitem{brock2018large}
Andrew Brock, Jeff Donahue, and Karen Simonyan.
\newblock Large scale gan training for high fidelity natural image synthesis.
\newblock In {\em ICLR}, 2018.

\bibitem{clouatre2019figr}
Louis Clou{\^a}tre and Marc Demers.
\newblock Figr: Few-shot image generation with reptile.
\newblock {\em arXiv preprint arXiv:1901.02199}, 2019.

\bibitem{dice1945measures}
Lee~R Dice.
\newblock Measures of the amount of ecologic association between species.
\newblock {\em Ecology}, 26(3):297--302, 1945.

\bibitem{fang2020identity}
Yuke Fang, Weihong Deng, Junping Du, and Jiani Hu.
\newblock Identity-aware cyclegan for face photo-sketch synthesis and
  recognition.
\newblock {\em Pattern Recognition}, 102:107249, 2020.

\bibitem{girshick2015fast}
Ross Girshick.
\newblock Fast r-cnn.
\newblock In {\em ICCV}, pages 1440--1448, 2015.

\bibitem{heusel2017gans}
Martin Heusel, Hubert Ramsauer, Thomas Unterthiner, Bernhard Nessler, and Sepp
  Hochreiter.
\newblock Gans trained by a two time-scale update rule converge to a local nash
  equilibrium.
\newblock {\em NeurIPS}, 30, 2017.

\bibitem{isola2017image}
Phillip Isola, Jun-Yan Zhu, Tinghui Zhou, and Alexei~A Efros.
\newblock Image-to-image translation with conditional adversarial networks.
\newblock In {\em CVPR}, pages 1125--1134, 2017.

\bibitem{karras2019style}
Tero Karras, Samuli Laine, and Timo Aila.
\newblock A style-based generator architecture for generative adversarial
  networks.
\newblock In {\em CVPR}, pages 4401--4410, 2019.

\bibitem{karras2020analyzing}
Tero Karras, Samuli Laine, Miika Aittala, Janne Hellsten, Jaakko Lehtinen, and
  Timo Aila.
\newblock Analyzing and improving the image quality of stylegan.
\newblock In {\em CVPR}, pages 8110--8119, 2020.

\bibitem{lee2018diverse}
Hsin-Ying Lee, Hung-Yu Tseng, Jia-Bin Huang, Maneesh Singh, and Ming-Hsuan
  Yang.
\newblock Diverse image-to-image translation via disentangled representations.
\newblock In {\em ECCV}, pages 35--51, 2018.

\bibitem{li2022Long}
Liang Li, Xingyu Gao, Jincan Deng, Yunbin Tu, Zheng-Jun Zha, and Qingming
  Huang.
\newblock Long short-term relation transformer with global gating for video
  captioning.
\newblock {\em TIP}, 2022.

\bibitem{li2016distributed}
Liang Li, Chenggang~Clarence Yan, Xing Chen, Chunjie Zhang, Jian Yin, Baochen
  Jiang, and Qingming Huang.
\newblock Distributed image understanding with semantic dictionary and semantic
  expansion.
\newblock {\em Neurocomputing}, 174:384--392, 2016.

\bibitem{li2020few}
Yijun Li, Richard Zhang, Jingwan Lu, and Eli Shechtman.
\newblock Few-shot image generation with elastic weight consolidation.
\newblock {\em arXiv preprint arXiv:2012.02780}, 2020.

\bibitem{liang2020dawson}
Weixin Liang, Zixuan Liu, and Can Liu.
\newblock Dawson: A domain adaptive few shot generation framework.
\newblock {\em arXiv preprint arXiv:2001.00576}, 2020.

\bibitem{lin2020tuigan}
Jianxin Lin, Yingxue Pang, Yingce Xia, Zhibo Chen, and Jiebo Luo.
\newblock Tuigan: Learning versatile image-to-image translation with two
  unpaired images.
\newblock In {\em ECCV}, pages 18--35. Springer, 2020.

\bibitem{liu2019few}
Ming-Yu Liu, Xun Huang, Arun Mallya, Tero Karras, Timo Aila, Jaakko Lehtinen,
  and Jan Kautz.
\newblock Few-shot unsupervised image-to-image translation.
\newblock In {\em ICCV}, pages 10551--10560, 2019.

\bibitem{liu2022unsupervised}
Zhenahuan Liu, Liang Li, Huajie Jiang, Xin Jin, Dandan Tu, Shuhui Wang, and
  Zheng-Jun Zha.
\newblock Unsupervised coherent video cartoonization with perceptual motion
  consistency.
\newblock In {\em AAAI}, 2022.

\bibitem{ma2018exemplar}
Liqian Ma, Xu Jia, Stamatios Georgoulis, Tinne Tuytelaars, and Luc Van~Gool.
\newblock Exemplar guided unsupervised image-to-image translation with semantic
  consistency.
\newblock In {\em ICLR}, 2018.

\bibitem{mo2020freeze}
Sangwoo Mo, Minsu Cho, and Jinwoo Shin.
\newblock Freeze the discriminator: a simple baseline for fine-tuning gans.
\newblock {\em arXiv preprint arXiv:2002.10964}, 2020.

\bibitem{noguchi2019image}
Atsuhiro Noguchi and Tatsuya Harada.
\newblock Image generation from small datasets via batch statistics adaptation.
\newblock In {\em ICCV}, pages 2750--2758, 2019.

\bibitem{ojha2021few}
Utkarsh Ojha, Yijun Li, Jingwan Lu, Alexei~A Efros, Yong~Jae Lee, Eli
  Shechtman, and Richard Zhang.
\newblock Few-shot image generation via cross-domain correspondence.
\newblock In {\em CVPR}, pages 10743--10752, 2021.

\bibitem{pang2021image}
Yingxue Pang, Jianxin Lin, Tao Qin, and Zhibo Chen.
\newblock Image-to-image translation: Methods and applications.
\newblock {\em arXiv preprint arXiv:2101.08629}, 2021.

\bibitem{robb2020few}
Esther Robb, Wen-Sheng Chu, Abhishek Kumar, and Jia-Bin Huang.
\newblock Few-shot adaptation of generative adversarial networks.
\newblock {\em arXiv preprint arXiv:2010.11943}, 2020.

\bibitem{saito2020coco}
Kuniaki Saito, Kate Saenko, and Ming-Yu Liu.
\newblock Coco-funit: Few-shot unsupervised image translation with a content
  conditioned style encoder.
\newblock In {\em ECCV}, pages 382--398. Springer, 2020.

\bibitem{salimans2016improved}
Tim Salimans, Ian Goodfellow, Wojciech Zaremba, Vicki Cheung, Alec Radford, and
  Xi Chen.
\newblock Improved techniques for training gans.
\newblock {\em NeurIPS}, 29:2234--2242, 2016.

\bibitem{tran2021data}
Ngoc-Trung Tran, Viet-Hung Tran, Ngoc-Bao Nguyen, Trung-Kien Nguyen, and
  Ngai-Man Cheung.
\newblock On data augmentation for gan training.
\newblock {\em TIP}, 30:1882--1897, 2021.

\bibitem{wang2008face}
Xiaogang Wang and Xiaoou Tang.
\newblock Face photo-sketch synthesis and recognition.
\newblock {\em PAMI}, 31(11):1955--1967, 2008.

\bibitem{wang2020minegan}
Yaxing Wang, Abel Gonzalez-Garcia, David Berga, Luis Herranz, Fahad~Shahbaz
  Khan, and Joost van~de Weijer.
\newblock Minegan: effective knowledge transfer from gans to target domains
  with few images.
\newblock In {\em CVPR}, pages 9332--9341, 2020.

\bibitem{wang2020semi}
Yaxing Wang, Salman Khan, Abel Gonzalez-Garcia, Joost van~de Weijer, and
  Fahad~Shahbaz Khan.
\newblock Semi-supervised learning for few-shot image-to-image translation.
\newblock In {\em CVPR}, pages 4453--4462, 2020.

\bibitem{wang2018transferring}
Yaxing Wang, Chenshen Wu, Luis Herranz, Joost van~de Weijer, Abel
  Gonzalez-Garcia, and Bogdan Raducanu.
\newblock Transferring gans: generating images from limited data.
\newblock In {\em ECCV}, pages 218--234, 2018.

\bibitem{xie2015holistically}
Saining Xie and Zhuowen Tu.
\newblock Holistically-nested edge detection.
\newblock In {\em ICCV}, pages 1395--1403, 2015.

\bibitem{yaniv2019face}
Jordan Yaniv, Yael Newman, and Ariel Shamir.
\newblock The face of art: landmark detection and geometric style in portraits.
\newblock {\em TOG}, 38(4):1--15, 2019.

\bibitem{yeh2017semantic}
Raymond~A Yeh, Chen Chen, Teck Yian~Lim, Alexander~G Schwing, Mark
  Hasegawa-Johnson, and Minh~N Do.
\newblock Semantic image inpainting with deep generative models.
\newblock In {\em CVPR}, pages 5485--5493, 2017.

\bibitem{yi2020unpaired}
Ran Yi, Yong-Jin Liu, Yu-Kun Lai, and Paul~L Rosin.
\newblock Unpaired portrait drawing generation via asymmetric cycle mapping.
\newblock In {\em CVPR}, pages 8217--8225, 2020.

\bibitem{yu2015lsun}
Fisher Yu, Ari Seff, Yinda Zhang, Shuran Song, Thomas Funkhouser, and Jianxiong
  Xiao.
\newblock Lsun: Construction of a large-scale image dataset using deep learning
  with humans in the loop.
\newblock {\em arXiv preprint arXiv:1506.03365}, 2015.

\bibitem{yu2018generative}
Jiahui Yu, Zhe Lin, Jimei Yang, Xiaohui Shen, Xin Lu, and Thomas~S Huang.
\newblock Generative image inpainting with contextual attention.
\newblock In {\em CVPR}, pages 5505--5514, 2018.

\bibitem{zhang2019consistency}
Han Zhang, Zizhao Zhang, Augustus Odena, and Honglak Lee.
\newblock Consistency regularization for generative adversarial networks.
\newblock In {\em ICLR}, 2019.

\bibitem{zhao2020differentiable}
Shengyu Zhao, Zhijian Liu, Ji Lin, Jun-Yan Zhu, and Song Han.
\newblock Differentiable augmentation for data-efficient gan training.
\newblock In {\em NeurIPS}, 2020.

\bibitem{zhao2020image}
Zhengli Zhao, Zizhao Zhang, Ting Chen, Sameer Singh, and Han Zhang.
\newblock Image augmentations for gan training.
\newblock {\em arXiv preprint arXiv:2006.02595}, 2020.

\bibitem{zhu2017unpaired}
Jun-Yan Zhu, Taesung Park, Phillip Isola, and Alexei~A Efros.
\newblock Unpaired image-to-image translation using cycle-consistent
  adversarial networks.
\newblock In {\em ICCV}, pages 2223--2232, 2017.

\end{thebibliography}
}

\end{document}


\title{Supplementary materials}

\author{First Author\\
Institution1\\
Institution1 address\\
{\tt\small firstauthor@i1.org}
\and
Second Author\\
Institution2\\
First line of institution2 address\\
{\tt\small secondauthor@i2.org}
}
\maketitle


\section{Experiment settings}
\label{sec:exp_set}

\noindent\textbf{Optimization Details.} We set initial learning rate $1.6e^{-4}$ and $1.8e^{-4}$ for the generator and discriminator respectively and adopt standard Adam optimizer for all of our experiments. We set batch size 4 in 5-shot and 10-shot settings and 1 in 1-shot setting. We stop training at 2500, 2000 and 1250 iterations in 10, 5 and 1-shot settings respectively. We operate on $256\times 256$ images for both the source and target domains except for the Cars$\rightarrow$ Wrecked Cars task because the source GAN of LSUN Cars is pre-trained at $512\times 512$ resolution. All of our experiments can be conducted on one NVIDIA RTX 3090 GPU with less than 1 hour.

\noindent\textbf{Choose of hyper-parameters.} For the disturbance correlation consistency loss $L_{dcc}$, we set the size of the radius of the neighborhood $r$ as $0.2\|z_i\|$, where $z_i$ is the center vector, so as to ensure that the corresponding neighborhood of $z_i$ is compact while all of the inner instances possess enough diversities. For the projector of latent space compression, we set the modulation coefficient $\alpha^l$ as $0$ for the $1^{st}\sim3^{rd}$ layers, $0.1$ for the $4^{th}\sim7^{th}$ layers, $0.3$ for the $8^{th}\sim9^{th}$ layers, $0.7$ for the $10^{th}\sim11^{th}$ layers and $0.9$ for the $12^{th}\sim14^{th}$ layers. The reason is that latent codes of different layers modulate the output images at distinct semantic levels in StyleGAN architecture. We set large $\alpha^l$ at the top layers and small ones at the bottom layers to alter the generated images' attributes at high semantic levels while maintaining the original spatial structure to a great extent.  

\noindent\textbf{Implementation details for spatial structural loss.} For the self-correlation consistency loss $L_{scc}$, direct computation of the self-correlation matrices is an $O((w\cdot h)^2)$ operation. To reduce memory requirements, we compute global self-correlation for feature maps at resolutions equal to or lower than $32\times 32$, but local self-correlation for feature maps at resolutions equal to or higher than $64\times 64$ by first adopting a $2\times$ average pooling and then splitting the feature maps into $16\times 16$ patches. For $L_{dcc}$, we only compute the pixel-wise mutual correlation between instances with the feature maps at resolutions equal to or lower than $128\times 128$. Calculating more feature maps with larger resolutions is fine if there is no memory limitation. We set the width of the slide window $\delta$ as $\frac{w}{4}$ where $w$ is the width of the feature maps.

\begin{figure*}[t]
  \centering
   \includegraphics[width=0.99\linewidth]{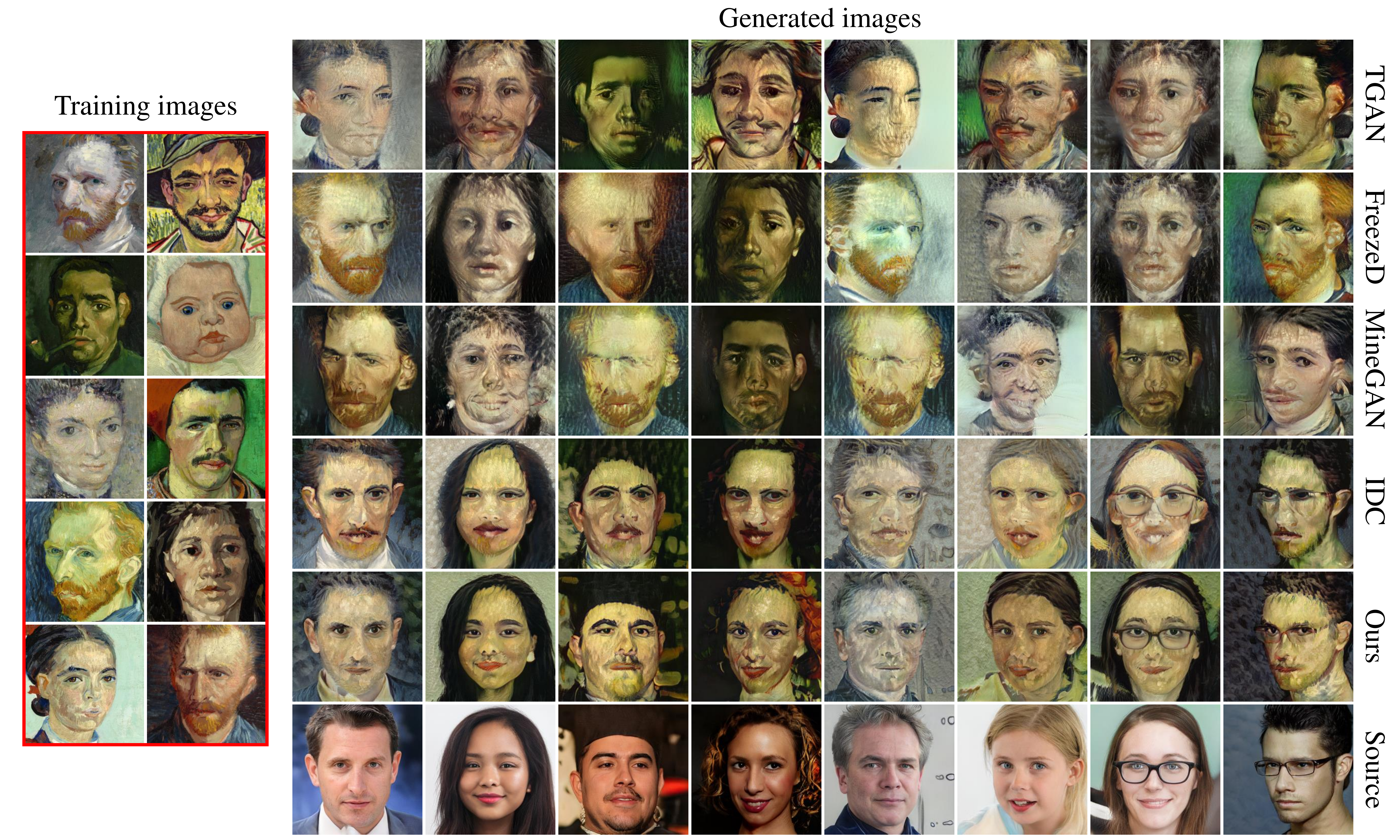}

   \caption{The visual comparison results on Faces$\rightarrow$ Van Gogh's face paintings in 10-shot setting.}
   \label{fig:face_van_10}
   \vspace{-3mm}
\end{figure*}
\begin{figure*}[t]
  \centering
   \includegraphics[width=0.99\linewidth]{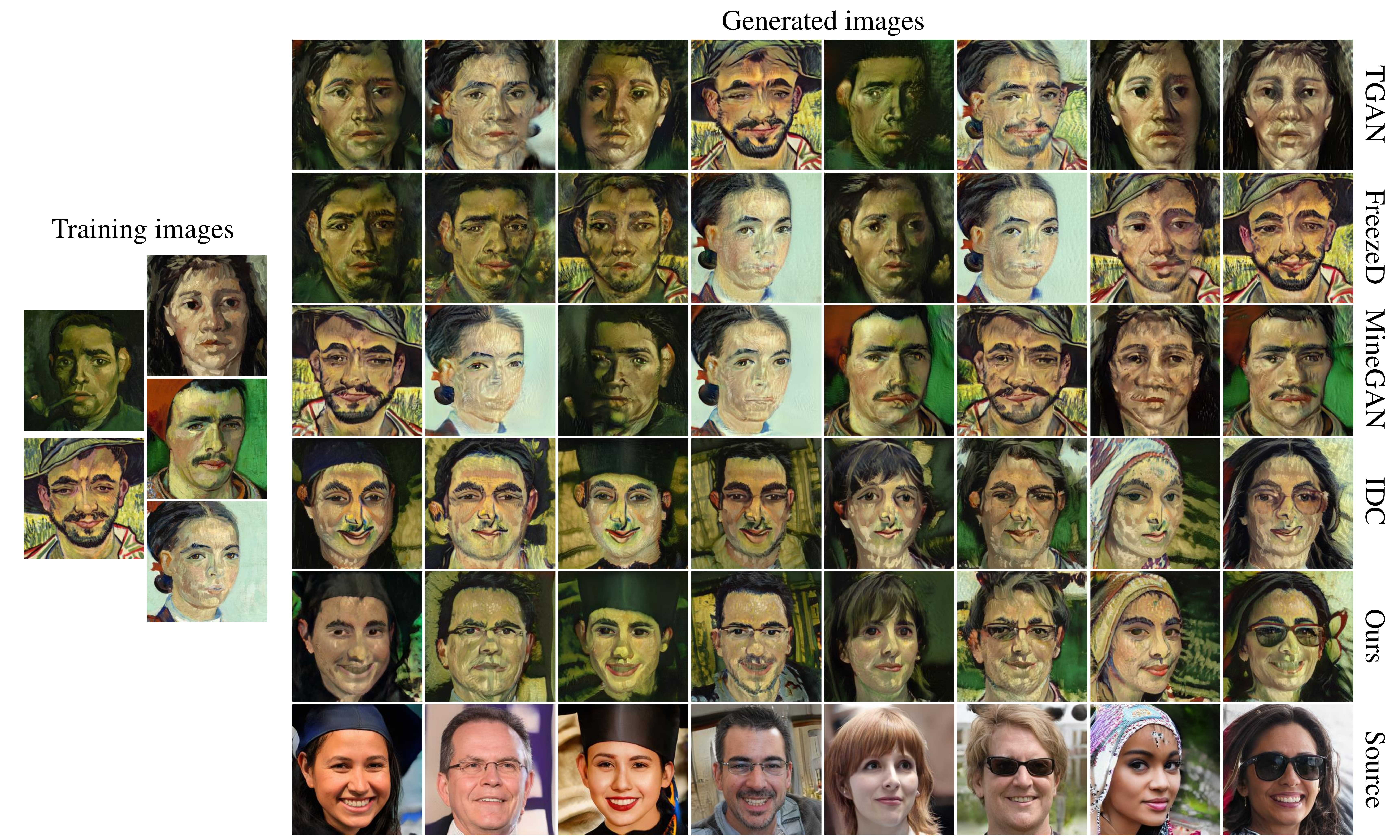}

   \caption{The visual comparison results on Faces$\rightarrow$ Van Gogh's face paintings in 5-shot setting.}
   \label{fig:face_van_5}
   \vspace{-5mm}
\end{figure*}

\section{Evaluation metric and overfitting}

In few shot setting, many fine-tuning based methods are prone to model overfitting, which means generators simply synthesis images similar with the few training samples. However, these overfitted models can get high IS in some cases, which verifies that IS can not deal with the problem of overfitting well. Table \ref{tab:quan_eval} (upper part) shows that TGAN, FreezeD and MineGAN obtain higher IS than IDC and our method on Faces$\rightarrow$ Van Gogh's paintings and Faces$\rightarrow$ Moise Kisling's paintings, but a higher IS does not mean a better generator. We show the visual comparison results on Faces$\rightarrow$ Van Gogh's paintings in Fig. \ref{fig:face_van_10} (10-shot) and Fig. \ref{fig:face_van_5} (5-shot). It is clear that our and IDC's generators do not suffer from the same overfitting problems as the generators of TGAN, FreezeD and MineGAN. Our method achieves the undisputed best visual results.

\begin{figure*}[t]
  \centering
  \vspace{-5mm}
   \includegraphics[width=0.99\linewidth]{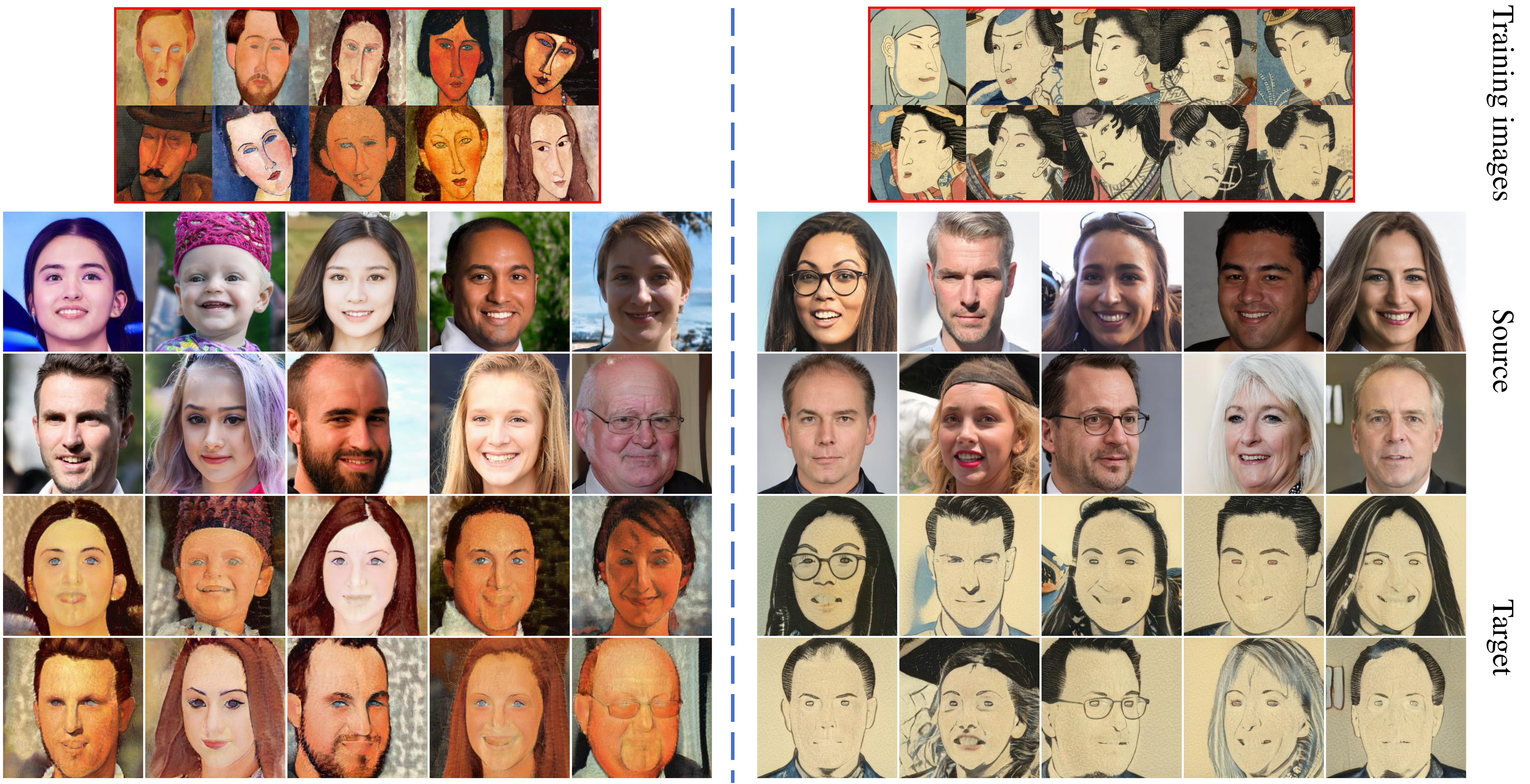}

   \caption{Results on Faces$\rightarrow$ Amedeo Modigliani's face paintings and Faces$\rightarrow$ Utagawa Kunisada's face paintings in 10-shot setting.}
   \label{fig:face_limitation}
   \vspace{-3mm}
\end{figure*}

Different from IS, the proposed SCS can evaluate the generated images from the perspective of spatial structure preservation capacity. SCS is very sensitive to model overfitting because an overfitted model can not generate images with structures similar to the source domain images. Table \ref{tab:quan_eval} (lower part) shows that our method and IDC obtain higher SCS than TGAN, FreezeD and MineGAN, which is consistent with the visual comparison results. Therefore, we claim that SCS, as a new metric, can effectively cover the shortcomings of IS in few shot GAN adaption scenarios.

\begin{table}[h]
\small
  \centering
  \renewcommand\arraystretch{1.2}
\resizebox{\linewidth}{!}{
\begin{tabular}{llllll}
\hline
\multicolumn{1}{l}{\multirow{2}{*}{Metric}} & \multicolumn{1}{l}{\multirow{2}{*}{Method}} & \multicolumn{2}{c}{f$\rightarrow$V} & \multicolumn{2}{c}{f$\rightarrow$M}  \\ \cline{3-6} 
  \multicolumn{1}{c}{}   &   \multicolumn{1}{c}{}     & {10-shot}            & {5-shot}           & {10-shot}            & {5-shot}           \\ \hline
\multirow{5}{*}{IS}     & TGAN                     & \textbf{2.39$^{\pm0.01}$ }    & 1.77$^{\pm0.05}$         & 2.05$^{\pm0.06}$        & \textbf{1.92$^{\pm0.06}$}   \\
                        & FreezeD        & 2.16$^{\pm0.02}$     & \textbf{1.81$^{\pm0.03}$ }        & 2.08$^{\pm0.03}$        & 1.90$^{\pm0.04}$    \\
                        & MineGAN             & 2.33$^{\pm0.03}$     & 1.74$^{\pm0.02}$         & 2.02$^{\pm0.05}$        & 1.88$^{\pm0.03}$   \\
                        & IDC            & 1.54$^{\pm0.01}$      & 1.40$^{\pm0.03}$  & 1.89$^{\pm0.03}$    & 
                        1.65$^{\pm0.05}$    \\
                        & Ours             & 1.76$^{\pm0.02}$  &1.56$^{\pm0.02}$ & \textbf{2.10$^{\pm0.04}$ } &  1.84$^{\pm0.05}$ \\ \hline
\multirow{5}{*}{SCS}    & TGAN         & 0.380              & 0.375            & 0.344       & 0.350           \\
                        & FreezeD                & 0.384              & 0.372            & 0.347       & 0.346      \\
                        & MineGAN               & 0.340              & 0.332            & 0.339       & 0.341        \\
                        & IDC           & 0.594             &0.568            & 0.524      &   0.494       \\
                        & Ours                     & \textbf{0.673}          &     \textbf{0.641}          & \textbf{0.639}        &  
                        \textbf{0.663} \\ \hline

\end{tabular}}

\small
  \centering
  \vspace{-2mm}
  \caption{Cope with model overfitting, IS VS. SCS. f represents faces source domain. V and M represent Van Gogh's paintings and Moise Kisling's paintings target domains. Best results are \textbf{bold}.}
  \label{tab:quan_eval}%
   \vspace{-3mm}
\end{table}

\section{Latent vectors interpolation}
\begin{figure}[h]
  \centering
   \includegraphics[width=0.99\linewidth]{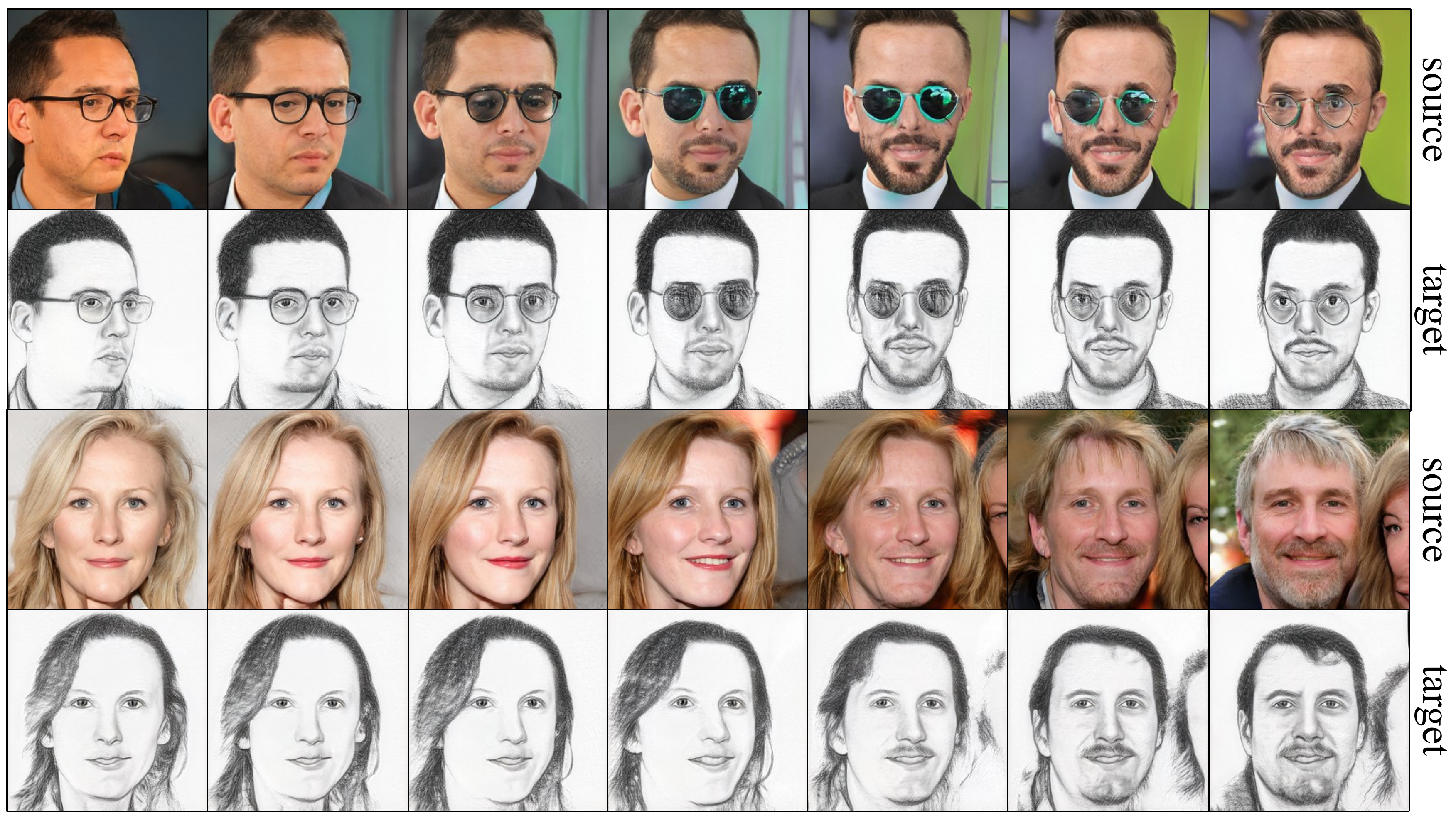}

   \caption{Interpolations between latent vectors. Source: Faces, Target: Sketches.}
   \label{fig:inter}
   \vspace{-3mm}
\end{figure}

The latent space is continuous rather than discrete, so we need to examine the smoothness of the generated image distribution through interpolation. Fig.~\ref{fig:inter} shows the results of interpolation between two latent vectors $z$ on Faces$\rightarrow$ Sketches in 10-shot setting. Although the amount of training data is small, a smooth and consistent interpolation is obtained by our method. We also show two GIF images to illustrate interpolation results of Faces$\rightarrow$Van Gogh, Moise Kisling, and Churches$\rightarrow$Van Gogh, Haunted Houses separately in 10-shot setting in the ZIP package.

\section{Limitations}

Although our proposed method is verified to be powerful in many few shot generative model adaption settings, there are still some limitations to be discussed here. The structural consistency constraint is not friendly to some target domains which are abstract or non-realistic. For example, paintings from Amedeo Modigliani generally have a surreal elongation of faces. Fig. \ref{fig:face_limitation} shows results of our method on Faces$\rightarrow$ Amedeo Modigliani's face paintings and Faces$\rightarrow$ Utagawa Kunisada's face paintings in 10-shot setting. It is a pity that the paintings of the target domains have unique characteristics in structure, such as elongated faces (left part) or angular faces (right part), but generated paintings of our method tend to preserve the structures of images from the source domain. 

\section{Additional visual results}

\begin{figure*}[t]
  \centering
   \includegraphics[width=0.99\linewidth]{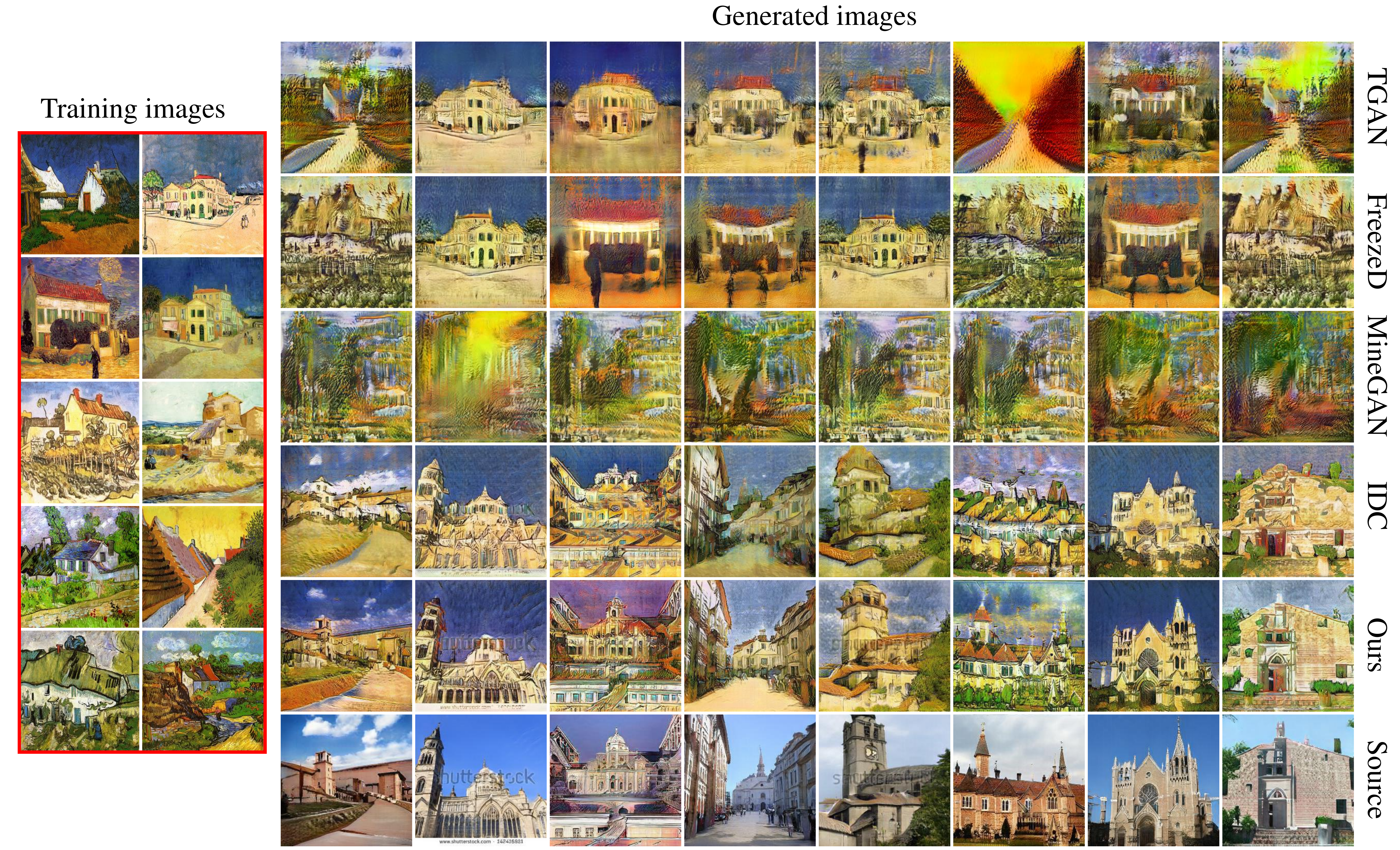}

   \caption{Comparison results with different methods in 10-shot setting. Source domain: Churches. Target domain: Van Gogh Village.}
   \label{fig:church_van_10}
   \vspace{-3mm}
\end{figure*}

\begin{figure*}[t]
  \centering
   \includegraphics[width=0.99\linewidth]{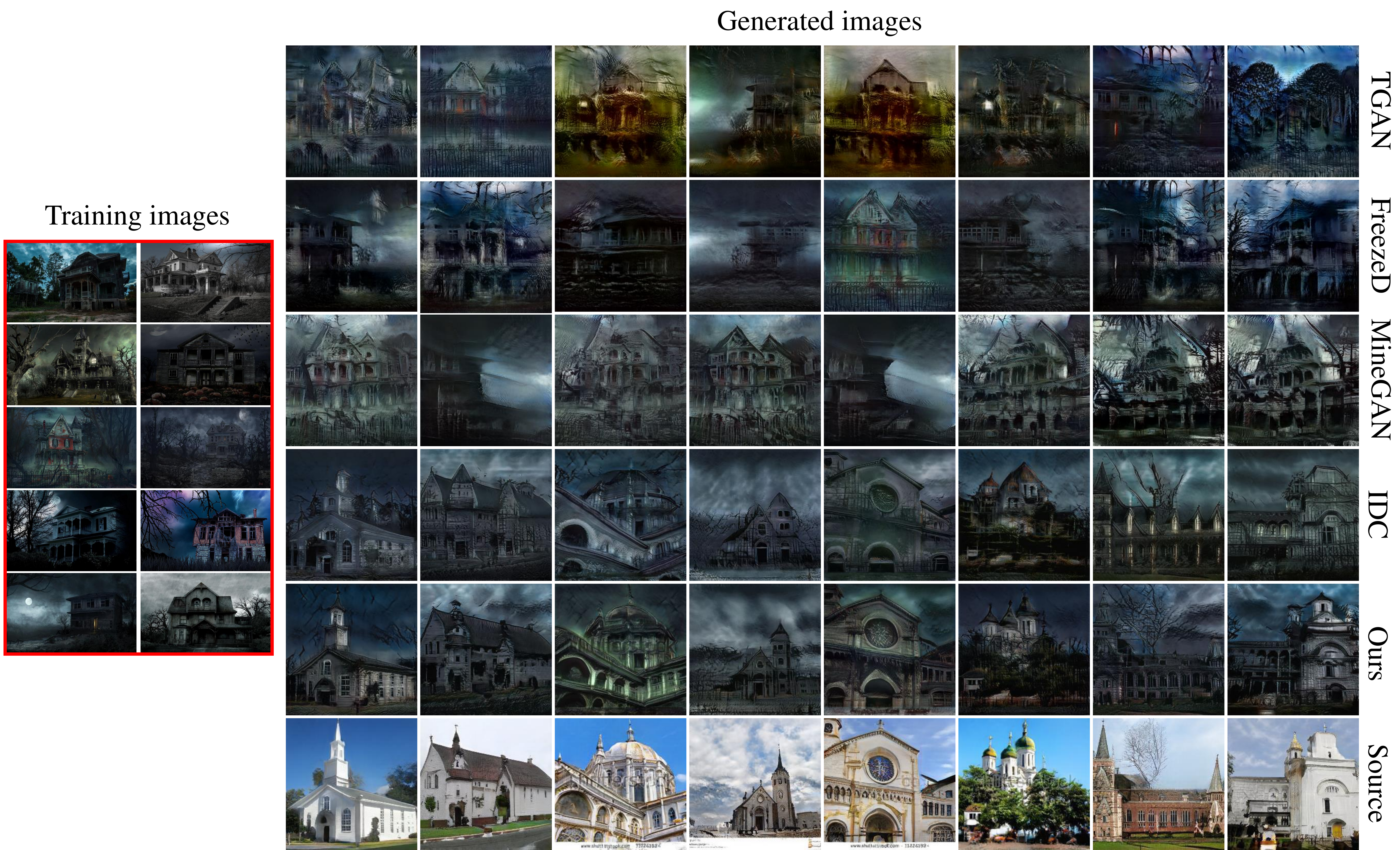}

   \caption{Comparison results with different methods in 10-shot setting. Source domain: Churches. Target domain: Haunted Houses.}
   \label{fig:church_haunted_10}
   \vspace{-3mm}
\end{figure*}

We've done extensive experiments on multiple source domain$\rightarrow$ target domain pairs. Due to page limitations, only results in some typical settings are included in the main paper. Fig.~\ref{fig:church_van_10} shows the comparison results with different methods on Churches$\rightarrow$ Van Gogh
Village in 10-shot setting. Fig.~\ref{fig:church_haunted_10} shows the comparison results with different methods on Churches$\rightarrow$ Haunted Houses in 10-shot setting. We can observe that TGAN, FreezeD and MineGAN overfit to the few training samples and synthesis images with poor visual diversity. As a competitive baseline method, IDC preserves some identities inherited from the source domains while we still witness many unnatural distortions and textures (e.g., vanishing or emerging windows and roofs). In comparison, our method mines the visual pattern of the training target samples and meanwhile preserves the cross domain structural consistency to a great extent.

